%% file: ACSAC-Paper.tex
\documentclass[sigconf]{acmart}
\usepackage{csquotes}
\usepackage{amsmath}
\usepackage{enumitem}
\usepackage{multirow}
\usepackage{float}
\usepackage{graphicx}
\usepackage{textcomp}
\usepackage{tikz}

\AtBeginDocument{%
  \providecommand\BibTeX{{%
    \normalfont B\kern-0.5em{\scshape i\kern-0.25em b}\kern-0.8em\TeX}}}

\DeclareMathOperator*{\argmax}{argmax}

\newcommand\detector{\textsf{Argos}}

\newcommand*\squared[1]{\tikz[baseline=(char.base)]{\node[shape=rectangle,fill,inner sep=1pt] (char) {\textcolor{white}{#1}};}}

\copyrightyear{2021}
\acmYear{2021}
\setcopyright{acmlicensed}\acmConference[ACSAC '21]{Annual Computer Security Applications Conference}{December 6--10, 2021}{Virtual Event, USA}
\acmBooktitle{Annual Computer Security Applications Conference (ACSAC '21), December 6--10, 2021, Virtual Event, USA}
\acmPrice{15.00}
\acmDOI{10.1145/3485832.3485904}
\acmISBN{978-1-4503-8579-4/21/12}

\acmBadgeR{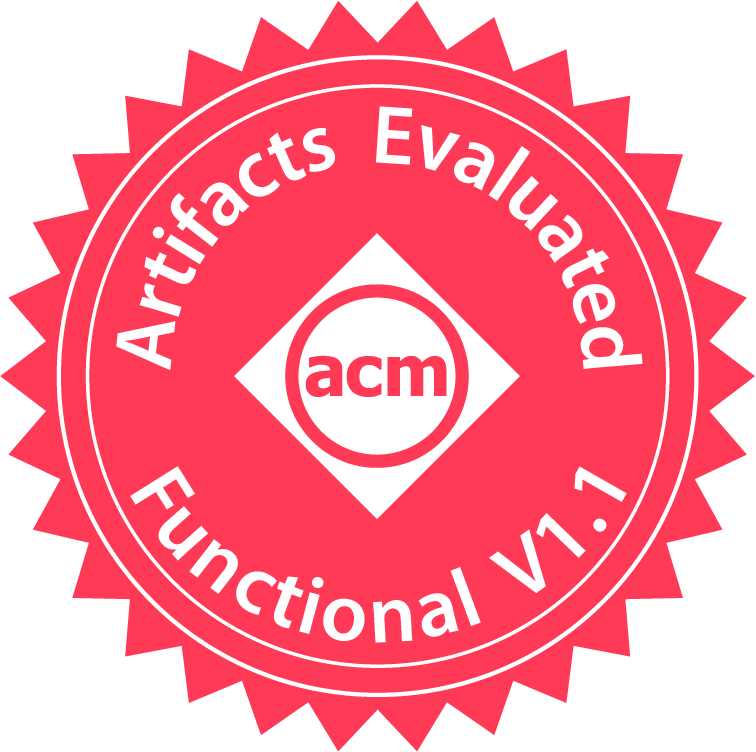}

\begin{document}

%\title{An Image of Two Souls: Adversarial Detection using Multi-view Inconsistency}
%\title{Adversarial Detection using Multi-view Inconsistency}
%\title{An Adversarial Image with Two Souls: Detecting Adversarial Examples using Multi-view Inconsistency}
%\title{An Adversarial Image with Two Souls: Towards Universal Adversarial Example Detection using Multi-view Inconsistency}
%\fancyhead{}
\title{Two Souls in an Adversarial Image: Towards Universal Adversarial Example Detection using Multi-view Inconsistency}
\renewcommand{\shorttitle}{Two Souls in an Adversarial Image: Towards Universal Adversarial Example Detection using Multi-view Inconsistency}

%\title{An Image of Two Souls: Detecting Adversarial Examples in Low Perturbation and White-Box Attacks}
%\title{An Image of Two Souls: A Multi-view Generative Approach for Adversarial Example Detection}

% \author{Sohaib Kiani}
% \affiliation{%
%     \institution{University of Kansas}
%     \country{}}
% \email{sohaib.kiani@ku.edu}
% \author{Sana Awan}
%     \affiliation{\institution{University of Kansas}
%     \country{}}
% \email{sanaawan@ku.edu}
% \author{Chao Lan}
% \affiliation{\institution{University of Oklahoma}}
% \email{clan@ou.edu}
% \author{Fengjun Li}
% \affiliation{\institution{University of Kansas}}
% \email{fli@ku.edu}
% \author{Bo Luo}
% \affiliation{\institution{University of Kansas}}
% \email{bluo@ku.edu}

%\author{Manuscript Submitted for Double-Blind Review}
% \email{ }
% \affiliation{
% \institution{\ }
% \city{\ }
% \country{}
% \postcode{}
% }

%\renewcommand{\shortauthors}{}

\author{Sohaib Kiani$^1$, Sana Awan$^1$, Chao Lan$^2$, Fengjun Li$^1$, Bo Luo$^1$}
\affiliation{%
    \institution{1. Department of Electrical Engineering and Computer Science, The University of Kansas, USA}
    \institution{2. School of Computer Science, The University of Oklahoma, USA}
    \country{}}
\email{sohaib.kiani@ku.edu, sanaawan@ku.edu, clan@ou.edu, fli@ku.edu, bluo@ku.edu}

%\author{Sohaib Kiani, Chao Lan, Fengjun Li, Bo Luo}
%\email{{sohaib.kiani,fli,bluo}@ku.edu; clan@ou.edu}
%\author{Fengjun Li}
%\email{fli@ku.edu}
%\author{Bo Luo}
%\email{bluo@ku.edu}
%\affiliation{%
%\institution{University of Kansas}
%\city{Lawrence}
%\state{Kansas}
%\country{USA}
%\postcode{66044}
%}
\renewcommand{\shortauthors}{Sohaib Kiani, Sana Awan, Chao Lan, Fengjun Li, Bo Luo}

% <concept_id>10010147.10010257.10010258.10010259.10010263</concept_id>
% <concept_desc>Computing methodologies~Supervised learning by classification</concept_desc>
% <concept_id>10002950.10003648.10003662</concept_id>
% <concept_desc>Mathematics of computing~Probabilistic inference problems</concept_desc>
% <concept_significance>300</concept_significance>

% \ccsdesc[500]{Computing methodologies~Supervised learning by classification}
% \ccsdesc[300]{Mathematics of computing~Probabilistic inference problems}




\begin{abstract}

{\color{black} 
In the evasion attacks against deep neural networks (DNN), the attacker generates adversarial instances that are visually indistinguishable from benign samples and sends them to the target DNN to trigger misclassifications. In this paper, we propose a novel multi-view adversarial image detector, namely \detector, based on a novel observation. That is, there exist two ``souls'' in an adversarial instance, i.e., the visually unchanged content, which corresponds to the true label, and the added invisible perturbation, which corresponds to the misclassified label. Such inconsistencies could be further amplified through an autoregressive generative approach that generates images with seed pixels selected from the original image, a selected label, and pixel distributions learned from the training data. The generated images (i.e., the ``views'') will deviate significantly from the original one if the label is adversarial, demonstrating inconsistencies that \detector~expects to detect. To this end, \detector~first amplifies the discrepancies between the visual content of an image and its misclassified label induced by the attack using a set of regeneration mechanisms and then identifies an image as adversarial if the reproduced views deviate to a preset degree. Our experimental results show that \detector~significantly outperforms two representative adversarial detectors in both detection accuracy and robustness against six well-known adversarial attacks. Code is available at: \url{https://github.com/sohaib730/Argos-Adversarial_Detection}
}

% Deep neural networks (DNNs) are proven successful and outperformed traditional machine learning methods in a wide range of tasks. Meanwhile, various cyber attacks against DNNs have been proposed in the literature, such as poisoning, evasion, backdoor, and model inversion. In the evasion attacks, the attacker generates adversarial examples that are visually indistinguishable with benign samples. Such adversarial examples are sent to the target DNN to trigger misclassifications. Several defense mechanisms have been developed in the research community. They are shown to be effective when the adversarial samples are generated with a higher amount of perturbation. However, it is still very challenging to detect well-crafted adversarial examples when the added perturbation is low. In this paper, we propose a generative approach that amplifies the inherent discrepancy between the visual content in the adversarial example and the misclassified label. We then attempt to identify adversarial samples by analyzing multiple generative views and measuring the inconsistency across the views. Experiment results show that the proposed approach significantly outperforms existing adversarial attack detectors, especially in the detection of low-perturbation attack samples.

%their detection accuracy drops when the added  decreases. 

\end{abstract}

% <concept_id>10010147.10010257.10010258.10010259.10010263</concept_id>
% <concept_desc>Computing methodologies~Supervised learning by classification</concept_desc>
% <concept_id>10002950.10003648.10003662</concept_id>
% <concept_desc>Mathematics of computing~Probabilistic inference problems</concept_desc>
% <concept_significance>300</concept_significance>

% \ccsdesc[500]{Computing methodologies~Supervised learning by classification}
% \ccsdesc[300]{Mathematics of computing~Probabilistic inference problems}

% \keywords{Adversarial Machine Learning, Auto-regressive Deep Generative Modeling, Multi-view classification}

\maketitle

\input{introduction}

\input{intuition}
\input{viewgeneration}

\input{experiment}

\input{discussions}

\input{relatedwork}

\input{conclusion}

\begin{acks}
The authors were supported in part by the US National Science Foundation under grant No.: IIS-2014552, IIS-2101936, DGE-1565570, DGE-1922649, and the Ripple University Blockchain Research Initiative. The authors
would like to thank the anonymous reviewers for their valuable comments and
suggestions. 
%We thank for the computational resources and technical help provided by ITTC, KU. 
%This project is funded by ... 
\end{acks}

\bibliographystyle{ACM-Reference-Format}
\bibliography{References}

%\vspace{4mm}
\appendix
\section{Adversarial Attacks against DNN}\label{sec:attacks}
Here we introduce the adversarial attack techniques that has been used to evaluate \detector~and the SOTA defense approaches. 

%\subsubsection{Fast Gradient Sign Method(FGSM)}

\vspace{2mm}
\noindent\textbf{Fast Gradient Sign Method (FGSM).} 
FGSM assumes the linear behavior in high-dimensional spaces is sufficient to generate adversarial inputs~\cite{FGSM}. Therefore, it constructs adversarial samples by applying a first-order approximation of the loss function. That is, given a data sample $(x,y)$ and the cross-entropy loss $L(\theta;x,y)$, an adversarial sample can be generated as:
\begin{equation*}
    x' =  x + \epsilon \cdot sign(\nabla_x L(\theta))
\end{equation*}
\noindent where $\nabla_x L(\theta)$ is the gradient of the loss function w.r.t. $x$.

\vspace{2mm}
\noindent\textbf{Projected Gradient Descent (PGD).} 
As an iterative version of FGSM, PGD~\cite{madry2018towards} selects the original image as a starting point and generates adversarial examples as follows. It can generate strong adversarial examples and thus is often used as a baseline attack to evaluate the defense designs. Procedurally, it starts from a benign image $x$ and iteratively modifies it into an adversarial image by
\begin{equation*}
\begin{aligned}
    x^{t+1} = clip_{x+\epsilon}[x^t + \alpha \cdot sign(\nabla_x L(\theta))]
\end{aligned}    
\end{equation*}
\noindent where $clip_{x+\epsilon}(x')$ is to keep $x'$ into the $\epsilon$-vicinity of $x$ and $\alpha$ can be set as $\epsilon/T$ with $T$ being the number of iterations.

\vspace{2mm}
\noindent\textbf{Momentum Iterative attacks.}
Inspired by the momentum optimizer, Dong et al.~\cite{dong2018boosting} proposed to integrate the momentum memory into the iterative process and derived a new iterative algorithm, called momentum iterative FGSM (MI-FGSM). Specifically, MI-FGSM updates the adversarial samples iteratively as follows.
\begin{equation*}
    \begin{aligned}
    x^0 &= x\\
    x^{t+1} &= clip_{x+\epsilon}[x^t + \alpha \cdot sign(g^{t+1})]
    \end{aligned}
\end{equation*}
\noindent where $g^{t+1} = g^t + \nabla L(\theta;x',y)/\left \|  \nabla L(\theta;x',y)\right \|_1$.
%\cite{dong2018boosting} also proposed a scheme to build an ensemble of models to attack a model in the black-box setting. 
%The basic idea is to consider the gradients of multiple models for the input and identify a gradient direction that is more likely to transfer to other models.

\vspace{2mm}
\noindent\textbf{DeepFool.}
DeepFool~\cite{moosavi2016deepfool} computes the minimal adversarial perturbation for an image based on an iterative linearization of the classifier. In particular, given a linear binary classifier, the minimal perturbation to change the classifier's decision for an input $x_0$ corresponds to the orthogonal projection of $x_0$ on $\mathscr{F}$, where $\mathscr{F}$ is the hyperplane at zero of $F$ (i.e., $\mathscr{F} =\{x: F(x)=0\}$). At each iteration $i$, the algorithm linearizes $F$ around the current point $x_i$ and computes the minimal perturbation of the classifier as:
\begin{equation*}
    \underset{r_i}{arg\,min}\left \| r_i \right \|_2 \;\;\;\; subject \, to \;\;\;\; F(x_i) + \nabla F(x_i)^T r_i = 0
\end{equation*}
\noindent where $r_i$ estimates the robustness of $F$ at $x_i$ and is computed as: 
\begin{equation*}
    r_*(x_i) = -\frac{F(x_i)}{\left \| \nabla F(x_i) \right \|_2^2} \nabla F(x_i)
\end{equation*}
The iteration stops when the sign of the classifier is changed. 
%DeepFool can be easily extended to multi-class classifiers, by treating them as multiple binary classifiers with the one-vs-all approach. 

\begin{comment}
It is given by the closed-form formula:
\begin{equation*}
r_*(x_0) := argmin  \left \| r \right \|_2\ \  s.t. sign(f(x_0 + r)) \neq sign(f(x_0))   
\end{equation*}
$r_*(x_0)$ can also be viewed as the robustness of the function $f$ at point $x_0$. 
\end{comment}

\vspace{2mm}
\noindent\textbf{The Carlini \& Wagner (C\&W) Attack.}
The C\&W attacks~\cite{carlini2017towards} are considered as optimization-based strong white-box attacks. Since, in our threat model we considered classifier Model $F(.)$ will be known for all type attacks, we can use C\&W to attack $F(.)$. In this method, the aim is to find the small perturbation $\delta$ satisfying:
\begin{equation*}
\underset{\delta}{arg\,min} \left \|\delta \right\|_p \;\;\; subject\,to \;\;\;\; F(x+\delta) = t,\ x+\delta \in [0,255]^n
\end{equation*}
\noindent where $t$ is the target label. To solve the optimization problem, an objective function $f$ is selected such that when $f(x+\delta) \leq 0$, $F(x+\delta) = t$. \cite{carlini2017towards} adopted the below objective function and generated adversarial samples with the $L_0,L_2$, and $L_{\infty}$ distance metrics.
\begin{equation*}
    f(x') = max(max\{Z(x')_i : i \neq t\} - Z(x')_t,-k)
\end{equation*}
\noindent where $Z$ is the logits function and $k$ is used to control the confidence of the misclassification.

\end{document}

%% file: introduction.tex
\section{Introduction}
\label{sec-intro}

With the growing computational power and the enormous data available from many sectors, applications with machine learning (ML) components are widely adopted in our everyday lives \cite{Sohaib_WOLF}. ML models especially deep neural networks (DNNs) now achieve human-level or even better performance in many challenging tasks, such as face/object recognition~\cite{sun2015deepid3,liang2015recurrent}, image classification~\cite{rawat2017deep}, autonomous driving \cite{grigorescu2020survey,muhammad2020deep}, and playing the game of Go~\cite{silver2016mastering}.
Meanwhile, a broad spectrum of cyber-attacks against DNNs such as poisoning \cite{steinhardt2017certified,shafahi2018poison,awan2021contra,tang2020demon}, evasion, backdoor \cite{pang2020tale,liu2018trojaning,tang2020embarrassingly}, and model inversion \cite{fredrikson2015model,wu2016methodology} has been proposed recently. One of the most harmful attacks is the evasion attack~\cite{madry2018towards,biggio2013evasion,nguyen2015deep,szegedy2013intriguing}, also known as adversarial examples, in which small imperceptible perturbations are added to input samples. These adversarial instances can fool the victim classifiers to make highly confident but erroneous predictions, and therefore have garnered significant attention from the defense side, e.g., \cite{song2018pixeldefend,metzen2017detecting,samangouei2018defensegan,feinman2017detecting,xu2017feature}.

\let\thefootnote\relax\footnotetext{The source code repository for this paper is located at \url{https://github.com/sohaib730/Argos-Adversarial_Detection}} 
%limits the applicability of ML models. For instance, ML models are vulnerable for the applications where adversary can be present such as malware detection, autonomous driving, and recommendation systems \cite{cyber,Cars, recomm}.

Existing countermeasures against the evasion attacks could be roughly divided into three categories: (i) to sanitize the input samples to (potentially) eliminate the adversarial perturbations~\cite{song2018pixeldefend,doan2020februus}; (ii) to enhance the robustness of the machine learning models, e.g., through adversarial training~\cite{madry2018towards,cao2017mitigating,cohen2019certified}; and (iii) to detect the adversarial perturbations and instances ~\cite{metzen2017detecting, feinman2017detecting,xu2017feature}. While these approaches have shown their effectiveness in detecting or preventing representative evasion attacks, 
%under the  settings, 
they still fall short in meeting the expectation for general and practical solutions. In particular, we have observed three key limitations in the state-of-the-art (SOTA) approaches. First, most of the current defenses are designed for black-box attacks, in which the attacker is assumed to know nothing about the defense mechanism (architecture and weights). However, in white-box attacks, the adversary with full knowledge of the victim model and the defense approach can easily generate adversarial examples that can evade the seemingly robust models or escape from detection~\cite{athalye2018obfuscated,carlini2017adversarial}. Second, defenses based on adversarial training achieve better performance against white-box attacks than the other two categories, but they incur an unavoidable penalty to clean model accuracy, as this line of solutions essentially trades classification accuracy for model robustness~\cite{tsipras2018robustness,tradeoff}. Moreover, the robustness of such models is still limited in certain cases. For example, for adversarial examples with high perturbations, %(e.g., $\epsilon>8$)
the accuracy of these approaches drops to 25\% or below~\cite{madry2018towards}. Finally, the SOTA attack detection solutions, especially the feature domain detectors, have shown promising performance in detecting adversarial examples with high perturbations, but many of them perform poorly in the detection of adversarial examples with low perturbations.
% Due to the difficulty of correctly classifying adversarial examples and the generality of
% their existence, other approaches focused on designing a detection scheme rather than increasing robustness of classifier models. In detection, the task is simplified to predict whether current sample is benign or adversarial. However, previously proposed detection schemes are again vulnerable against adversarial examples[ten det]. The underlying reason is if an attacker can fool a classifier then they can fool a detector as well because in reality it's a binary classifier. Therefore, in recent past technical community haven't really focused on building robust ML detectors. However, in our work we will show that it's possible to build robust detectors that can guarantee safety against all type of attacks.

In this paper, we present the \detector~solution, which attempts to tackle the challenge of adversarial example detection from a novel perspective. Intuitively, we have observed that an adversarial example, regardless of the level of perturbation, contains two inherently contradictory ``souls'': (i) the image content that is visually indistinguishable from benign samples, and (ii) the added invisible perturbation that corresponds to an adversarial label. When a victim classifier is misled by the adversarial perturbation, it generates a wrong label for the input adversary instance. We argue that the inherent discrepancy between the visible ``clean'' content and the misclassified label, which is forced by the invisible adversarial content, could be utilized in detecting adversarial examples. To amplify the discrepancy between the two ``souls'' to an extent that can be detected, we propose a novel {\em generation-and-detection} approach that first adopts an autoregressive generative model to regenerate images (called {\em views} in this work) using different sets of seed pixels from the original image and the predicted label (Section~\ref{sec:viewgen}), and then detects {the \textit{inconsistencies} between the original image and the view ensembles as well as across the views}  with a multi-view detector named \detector~(Section~\ref{sec:detection}).

%Multiple views are generated with different sets of seed pixels, such that the views would demonstrate significant inconsistency and many, if not all, of them would be quite different from the input image.

The rationale behind this design is two-fold. First, for a clean input, all the views generated with different seeds and the same correct label should be visually and distributionally similar so that the inconsistencies across the views and the original input are very low. On the contrary, a successful adversarial instance that fools the neural network classifier 
will result in a misclassified label and inevitably cause inconsistencies between the original image and the generated views as we expect. These views are not only significantly different from the original input, but more importantly, different from each other. Therefore, we further develop a novel model to measure and evaluate such inconsistencies for adversarial sample detection. Another advantage of the \detector~approach is that it does not rely on any prior knowledge nor make any assumption about the adversarial attack method (as long as the attack is successful). As a result, it can effectively detect unknown attacks. %We develop the first prototype of this adversarial example detector, which employs an autoregressive generative approach for multi-view generation and then examines the consistencies for adversarial image detection. 

Our contributions are summarized as follows.
\vspace{-0.5mm}
\begin{itemize}[leftmargin=*]
\item We are among the first to identify the inherent discrepancy between the (visually) benign content and the misclassified label in a successful adversarial example attack and leverage such discrepancy for attack detection. %We argue that the discrepancy could be amplified and further utilized in the detection of adversarial examples. 
\item We propose a novel generation-and-detection approach to amplify the imperceptible perturbation with an autoregressive generative model and develop a multi-view detector to measure the (in)consistencies across the generated images for adversarial example detection. %constructing synthetic images using both the image content and the adversarial label as seeds. We further develop a novel multi-view approach that measures the (in)consistencies in the generated images for adversarial example detection. 

\item Through extensive experiments, we show that the \detector~detector is highly accurate and robust. Moreover, it significantly outperforms SOTA solutions against the strongest adversarial attacks.
%\item Experiments with multiple datasets and attack methods demonstrate that 
\end{itemize}

The rest of the paper is organized as follows: We present the threat model in Section~\ref{sec:attackmodel} and an overview of our solution in Section~\ref{sec:solutionoverview}. We present the detailed design of \detector~in Section~\ref{sec:detector}, followed by the experimental results and some discussions in Sections~\ref{sec:exp} and \ref{sec:dis}. Finally, we briefly summarize the related work in Section~\ref{sec:rel} and conclude our work in Section~\ref{sec:con}.

%% file: intuition.tex
\section{Background and Threat Model}
\label{sec:attackmodel}

\subsection{Basic Notations and Objective}
\label{Attack_Model}

%The proposed adversarial detector aims to tackle the evasion attacks against deep neural networks (DNNs), a.k.a. adversarial examples.

%For any clean sample $(x,y)$, whose classified output is $F(x)=y$, the corresponding adversarial example $x'$ gives $F(x') \neq y$ and 

%is within   as given by Goodfellow et al. \cite{goodfellow2014explaining}. Also, we will specify $\epsilon$ in terms of pixel values in the range [0, 255]. 

We denote benign image as $x$, its true label as $y$, and the corresponding adversarially perturbed image as $x'$. We denote input image for classification/detection as $z$, which can be benign or adversarial i.e. $z=\{x,x'\}$. 
We assume that all images have been flatten into
$n$-dimensional vectors, 
i.e., $ z \in \mathbb{R}^{n}$. 
Let $F : \mathbb{R}^{n} \rightarrow \mathbb{R}$ 
be a deep neural network (DNN) classifier, mapping 
an input image $z$ to a probability vector $f(z)$, such that predicted label $F(z) = \argmax(f(z))$.
Let $h(z)$ be the vector presentation of 
$z$ at the penultimate layer of the network.  
We assume $F$ is standardly optimized from 
a labeled training set using a loss function
$L(\theta)$, where $\theta$ is the set 
of network weights (to optimize). 

Recall $F$ is expected to correctly classify images. Under an attack with $x'$, $F$ may give wrong prediction, i.e., $F(x') \neq y$. Our goal is to detect $x'$ when it launches a successful attack. We propose to build a  detector 
$D : \mathbb{R}^{n} \times \mathbb{N} \rightarrow \{0, 1 \}$, mapping from an image and its predicted label to a 
binary variable indicating whether the image is adversarial. If an image $z$ receives $D(z, F(z)) = 1$, then it is likely 
an adversarial instance and its predicted label $F(z)$ should be voided or at least alerted. 

\begin{figure*}[!ht]
  \centering
  \includegraphics[width=\textwidth]{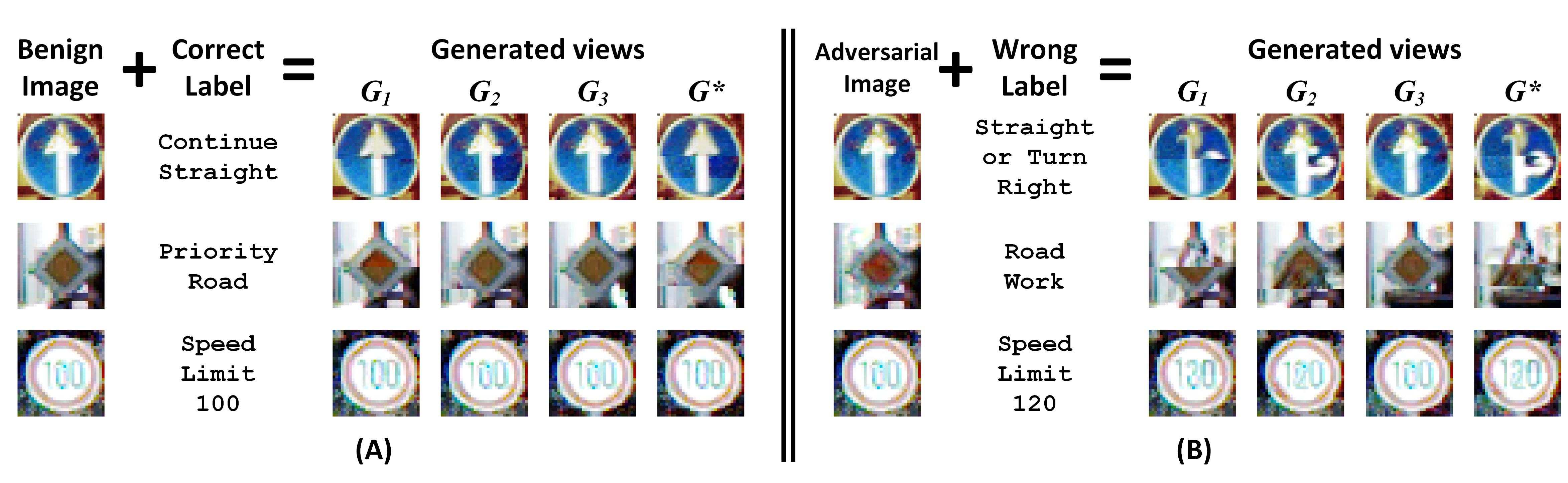}
  \vspace{-2mm}
  \caption{Examples of generated views for benign and adversarial images: each image receives a label from the classifier, four views are generated from the seed image and the label. (A) Benign images; (B) Adversarial images. }
  \label{fig:gen_images}%\vspace{-1mm}
\end{figure*}

\subsection{The Threat Model}
In this paper, we follow the standard threat model of evasion attacks against deep neural networks. The adversary, acting as a legitimate user of the victim DNN, deliberately introduces adversarial perturbation to $x$ and generates the corresponding $x'$, which is visually indistinguishable from $x$. 
%The adversary, acting as a legitimate user of the victim DNN, deliberately introduces adversarial perturbation to the clean image $x$ and generates the corresponding adversarial example $x'$, which is visually indistinguishable from $x$. 
The adversarial example is supposed to be within 
a certain $L_p$ distance from $x$, i.e., 
$\left \| x-x' \right \|_p \leq \epsilon$, 
where $\epsilon\in[0,255]$ is called the \textit{perturbation budget}. 

As introduced above, the victim system has a DNN classifier $F$ and an optional adversarial input detector $D$. Ultimately, the adversary has two objectives that are orthogonal: (Obj-1) to fool the victim DNN to misclassify $x'$ to a wrong label 
(i.e., $F(x')\neq y$); and (Obj-2) to fool the adversarial input detector (if it exists) to classify $x'$  as benign, i.e., $D(x', F(x'))$=0. 

Prior works in the literature %on adversarial examples 
consider two threat models, the black-box (Limited knowledge) and the white-box (Complete knowledge) settings \cite{carlini2017adversarial}. Both settings assume the attacker has full knowledge of the victim DNN $F(\cdot)$, including all the model parameters/weights, to craft effective adversarial instances. However, the attackers have different knowledge about the detector:
%A successful adversarial example is the one that fool both classifier and in place defense scheme\cite{carlini2017adversarial}, which is a detector in our case. Following are the possible two attack scenarios determined by the attacker's knowledge:

\begin{itemize}[topsep=0mm, wide, labelindent=0pt]
\item{\bf {\em The black-box setting}} assumes the attacker has full access to $F(.)$, but does not know anything about the (existence of) detector $D(.)$. In this case, the attacker only attempts to achieve Obj-1. 
\item{\bf {\em The white-box setting}} assumes the attacker has full knowledge of both the victim DNN and the detector, including all the weights, but not the testing-time randomness. In this case, the attacker attempts to achieve both Obj-1 and Obj-2, as defined above. 
\end{itemize}
%We have not considered oblivious attack setting, where attacker has no knowledge about classification and detection network. Because, it's safe to assume that if the detection scheme is working in Limited knowledge setting then it will work in oblivious setting as well.

From the defense perspective, the detector is expected to work in a semi-autonomous setting, where it only interacts with the DNN owner when an adversarial instance is detected. It has full knowledge of the victim DNN and full access to the training data. We do not make any assumption on the attack method that will be employed to generate adversarial examples. To develop such a general-purpose detector, any assumption made in the design should be \textit{generalizable} to unknown attacks and the defense mechanism should not rely on a large number of adversarial examples from any specific attack(s). Last, a benign input may be mistakenly classified into a wrong label by the victim DNN, since DNNs cannot achieve 100\% accuracy. The misclassified benign inputs are considered outside of the scope of the detector. We briefly compare and discuss the adversarially misclassified samples and naturally misclassified samples in Section \ref{sec:dis}.

% it asks user to intervene or invalidate the classified label when input image is identified as adversarial. 
% The detectors input vary based on detection scheme. In our design we use, input image and classified label to perform detection i.e. .

%the defender is expected to make no assumption of the attack

%is not designed or optimized to handle only one or a few attacks. The

%no large amount of adversarial data to train .....

\vspace{1mm}\noindent\textbf{Adversarial Attacks.} In this paper, we will evaluate \detector~and other adversarial image detectors with the following attacks: Fast Gradient Sign (FGSM) \cite{FGSM}, Projected Gradient Descent (PGD) \cite{madry2018towards}, Momentum Iterative Method (MIM) \cite{dong2018boosting}, DeepFool \cite{moosavi2016deepfool}, the Carlini \& Wagner (C\&W) Attack \cite{carlini2017towards}, and a white-box attack designed by us (to be elaborated in Section \ref{sec:exp}). A brief introduction to each of the existing attacks could be found in Appendix \ref{sec:attacks}. For more details of the attacks, please refer to their original publications.

\section{Solution Overview}\label{sec:solutionoverview}
Intuitively, an adversarial image contains two ``souls'': (i) the visible content, which is visually consistent with the clean label; and (ii) the adversarial perturbation, which is invisible but pushes the victim DNN to generate a wrong label. These two ``souls'' are fundamentally discrepant, while the ``stronger'' soul will win the competition at the DNN. For a well-crafted adversarial image, a very small perturbation is sufficient to trick the victim DNN to classify it into a wrong label. Conventional adversarial detectors use the added perturbation as a source of information to identify adversarial examples. They fall short in detecting low-perturbation attacks because the added noise is almost imperceptible or against the white box adversaries who control the added perturbation to fool the detector. 

The core idea of \detector~is to amplify the discrepancy between the visual content of the image and the wrong label that is forced by the invisible perturbation. In particular, we propose to generate different views from a part of the original image (seed pixels) and the predicted label using an autoregressive generative approach. Next, we present two sets of examples that inspire our design. %Figure \ref{fig:gen_images} demonstrates examples of this generative approach. 

\vspace{1mm}\noindent\textbf{Observations.} In Figure \ref{fig:gen_images} (A), benign images from the GTSRB dataset \cite{stallkamp2012man} are correctly classified by a deep neural network. We invoke a generative model, namely PixelCNN++~\cite{salimans2017pixelcnn++}, to produce four views ($G_1$ to $G^*$) from the source images and their predicted labels. Each view ($G_1$ to $G_3$) was generated using a different number of seed pixels. $G^*$ is an integration of the three views (we will discuss the details of view generation in Section~\ref{sec:detector}). As shown in the figure, all four views appear to be similar and visually consistent with the source image. In Figure~\ref{fig:gen_images} (B), three adversarial images, which look identical to the benign ones, are sent to the same DNN and receive adversarial labels. We generate four views from the adversarial images and the corresponding misclassified labels. From the figure, we have the following observations: (i) the generated images appear very inconsistent, (ii) both the original image and the adversarial label (i.e., two ``souls'') are somewhat reflected in most of the generated images, and (iii) most of the generated images visually deviate from the original image. 

\vspace{1mm}\noindent\textbf{\detector~Overview.} With the observations, we design a multi-view adversarial image detector, namely \detector.  As shown in Figure~\ref{fig:overview}, the end-to-end image classification and adversarial example detection approach consist of four steps: (1) An input image $z$, benign or adversarial, is sent to the target DNN, which assigns a label $F(z)$ to the input. We do not know whether the image is clean or adversarial, or whether the label is correct or wrong. (2) Both $z$ and $F(z)$ are sent to a generative model, which creates multiple views ($G_1, G_2, G_3$ and $G^*$) using the label $F(z)$ and different portions of $z$ as seeds. In each view $G_i$ in the figure, the region of the image highlighted in the red dashed rectangle is generated. (3) All the views, as well as the original image $z$, are sent to the decision module, which measures the discrepancies among them to determine if the label is benign or adversarial. (4) Finally, based on the output from \detector, we choose to accept the label, or reject the label and raise an alert. 

\color{black}\vspace{1mm}\noindent\textbf{Advantages of \detector.} \detector~does not make any assumption of the attack method, except the fact that the attacker switches the output label of the adversarial example ($x'$) from $y$ to $t$, while $x'$ is still visually similar to other samples in class $y$. The view generation mechanism has a single objective, which is to amplify the inconsistency, if any, among the visual content of the image and the predicted label. Therefore, \detector~is designed to detect all adversarial examples without distinction. It does not rely on a specific feature or algorithm of the attacks or a specific type or strength of the added perturbation, so that \detector~is highly generalizable to known or unknown attacks. It is also more robust against white-box attacks than other approaches (to be demonstrated in Section~\ref{sec:exp}). In addition to adversarial detection, \detector~can also identify some misclassified benign samples (to be discussed in Section~\ref{sec:dis}). 

% The proposed multi-view approach can be combined with conventional single view detection approaches to further boost the performance. Since, multiple views provides complimentary information for detection that can not be extracted with single view.

\color{black}

%% file: viewgeneration.tex
\section{\detector: the Multi-view Detector for Adversarial Examples}
\label{sec:detector}

%In this section, we first briefly revisit the co

% In this section, we briefly revisit conditional pixelCNN and present the technical details of the
% \detector~approach, including the view generator and the respective detector. 

\subsection{A Revisit of Conditional PixelCNN}

Let $z=(z_1,...z_n)$ be an $n$-dimensional 
vector representation of a flattened image $z$, 
where all pixels are assumed sorted in the raster
scan order, i.e., sorted row by row (from top 
to bottom) and pixel by pixel within each row 
(from left to right).

PixelCNN ~\cite{van2016pixel} is an autoregressive  modelling technique that models the joint pixel
distribution of $z$ by factoring it into a product of the following conditional pixel distributions: 
\begin{equation*}
p(z) = \prod_{i=1}^{n}p(z_{i}|z_{1},....z_{i-1}).  
\end{equation*}
The right distributions are modeled by 
a shared convolutional neural network 
learned from a set of training images. 

PixelCNN is widely applied to generate diverse
images that are capable of capturing the 
high-level structure of the training images~\cite{NADE}. In practice, it generates 
an image by selecting its first $k$ pixels 
(called seeds) and generating the rest one 
by one, with the $i_{th}$ pixel randomly 
sampled from 
\begin{equation}
p(z_{i} \mid z_{1}, \ldots, z_{i-1}).     
\end{equation}

Conditional PixelCNN~\cite{oord2016conditional,salimans2017pixelcnn++} is an enhanced version of 
pixelCNN which further conditions each pixel
distribution on a given label $\hat{y}$ (not
necessarily $y$), with an aim of improving the visual quality of the generated images, i.e., it models 
\begin{equation}
\label{eq:conditionalpixelcnn}
p(z \mid \hat{y}) = \prod_{i = 1}^{n}\, p(z_{i}|z_{1},....z_{i-1}, \hat{y}).  
\end{equation}

In this paper, we discover a new application 
of the above label dependence for detecting 
adversarial image $x'$. By setting $\hat{y} 
= F(x')$, we show (\ref{eq:conditionalpixelcnn}) 
can be applied to generate low-quality copies 
of $x'$, based on the inconsistency between
its visually unchanged content (i.e., 
$||x - x'||_{p} \leq \epsilon$) and 
mis-classified label (i.e., $F(x') \neq y$). 
We hypothesize that such inconsistency can 
be leveraged to detect adversarial images.

\subsection{View Generation and Ensemble}
\label{sec:viewgen}

Given a test image $z$, we first generate a copy of it, denoted by $G(z,F(z))$, using conditional pixelCNN based on both $z$ and $F(z)$. Intuitively, if $z$ is a successful adversary, it should visually look like one object but be classified as another. 
In the experiment, we observe such inconsistency being amplified in the generated copies. In this paper, we have denoted generated image as $G$ and omitted it's input $(z,F(z))$ to save space.

Let us re-visit Figure \ref{fig:gen_images} that shows two sets of results: 
in (A), the generated copies of three benign images are conditioned on correctly predicted labels. 
In each row, the left-most column shows the test image and the following columns show the generated copies 
using different sets of seeds. 
And in (B), the generated copies of three adversarial images are conditioned on misclassified labels. 

Take the top image in (B) as an example. 
The input image is undoubtedly a `go-straight' 
sign by appearance, but is actually an 
adversarial image misclassified 
as `go-straight-or-turn-right'. 
Then, we see the first generated copy 
$G_{1}$ looks like an integration of go-straight and go-straight-or-turn-right, 
which deviates significantly from the input image. 
While similar observations can be found for the other two adversarial examples 
(the `Road Work' sign in the second row 
and the `Speed Limit 120' sign in the third row), 
we do not see such deviation in benign images 
in (A). This implies that the inconsistencies between the visual features and the misclassified labels can 
be re-captured and used to detect adversarial images. 
\begin{figure}
  \centering
  \includegraphics[width=\columnwidth]{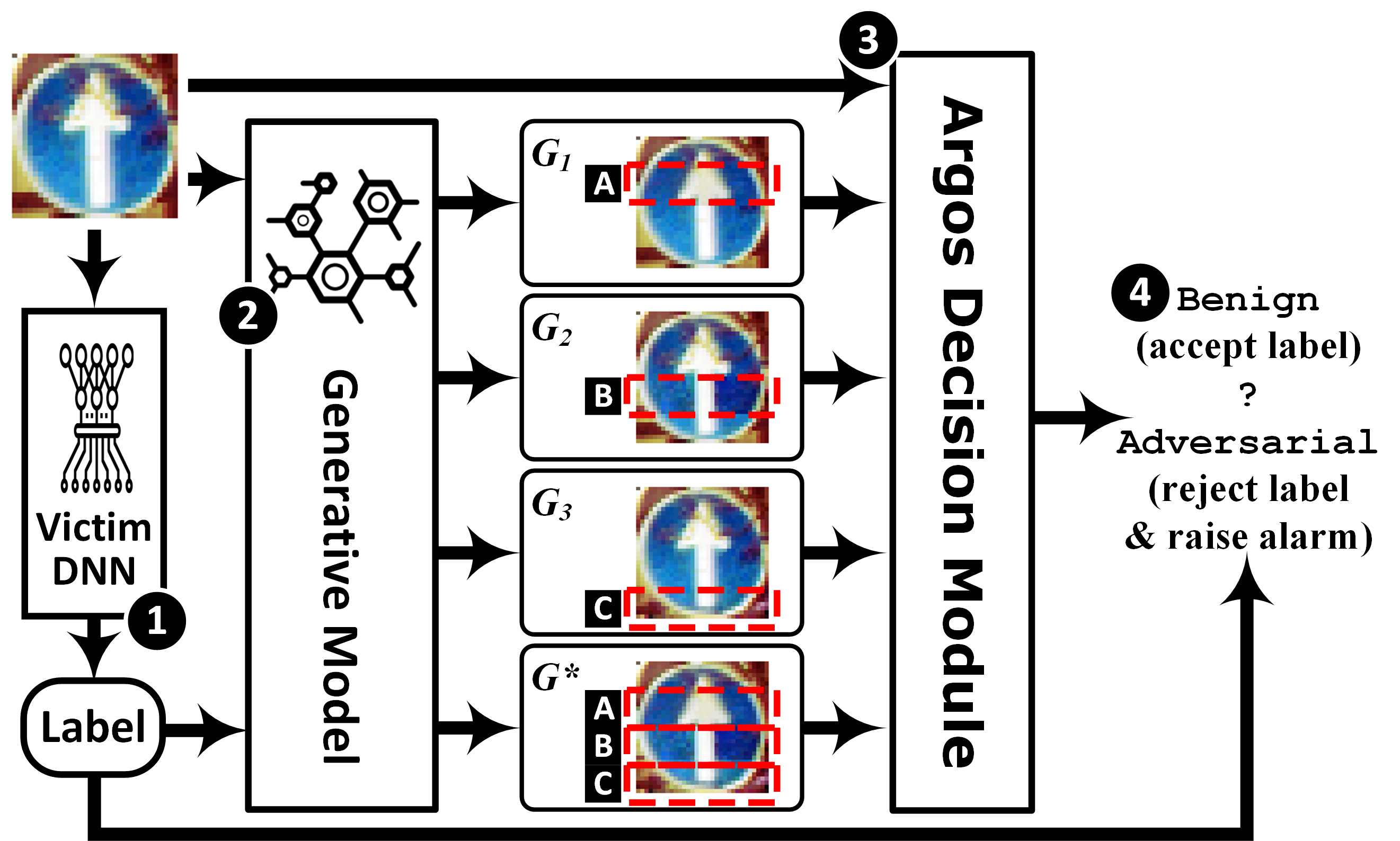}
  \caption{An Overview of the \detector~approach. }
  \label{fig:overview}
\end{figure}
To generate a copy $G$ of a test image $z$, a naive way is to select the first 
$x$ pixels (horizontally or vertically) of $z$ and generate the rest using Eq.  (\ref{eq:conditionalpixelcnn}). However, there could be many different approaches to select the seeds and the generated portion of the image. The view generation strategy in \detector~is mostly empirical. We have observed that a top-down approach generates better views for benign images. Meanwhile, we also observed that the quality of the generated rows of pixels will decrease when the new rows are far away from the seeds,  as their sampling distributions are conditioned heavily on the previously generated pixels, instead of the observed seeds. 
This may add noise even for benign images and consequently lower the detection accuracy. 
To mitigate this issue, we only generate a small portion of the image in each copy $G$, and adopt the remaining pixels from the original input image. Since we do not know where exactly the object is located in the image, we generate multiple copies of $z$, each called 
a \textit{view} ($G_k$) with each focusing on a different area of the image. To capture the overall effect, the final view $G^*$ is assembled from all the generated pixels from previously generated views.

In \detector, the $k_{th}$ view $G_{k}$ is created by taking the first $m_{k}$ rows of pixels in $z$ as seeds to generate rows ranging from $(m_{k}+1,m_{k+1})$.  For an image with $n$ number of rows we set $m_{k}$ = $k \times \frac{n}{4}$. Thus, we generate a quarter of an image in each view which gives three $G_k$ views and the final view $G^*$ is assembled from the generated rows from all $G_k$ views. Among all the views, the area that was not filled by generated rows will be copied as is from the input image $z$. For instance, with an input image of 32$\times$32 pixels, we have $m_1$=8, $m_2$=16, and $m_3$=24. As shown in Figure \ref{fig:overview}, in view $G_1$, rows 9 to 16 are generated with Eq.  (\ref{eq:conditionalpixelcnn}) (labeled as {\footnotesize \squared{\texttt{\textbf{A}}}} in Figure \ref{fig:overview}), while all other rows are directly adopted from $z$. In the same way, rows 17 to 24 are generated in $G_2$ ({\footnotesize \squared{\texttt{\textbf{B}}}} in Figure \ref{fig:overview}), and rows 25 to 32 are generated in $G_3$ ({\footnotesize \squared{\texttt{\textbf{C}}}} in Figure \ref{fig:overview}). Finally, view $G^*$ adopts the generated regions from $G_1$ to $G_3$ %({\footnotesize \squared{\texttt{\textbf{A}}}}, {\footnotesize \squared{\texttt{\textbf{B}}}} and {\footnotesize \squared{\texttt{\textbf{C}}}} in Figure \ref{fig:overview}) 
to generate an integrated view. 

%We will further discuss view generation strategies in Section \ref{sec:dis}. 

In practice, $G_k$ views may be generated 
in parallel to improve the computational efficiency of \detector. 
More efficient implementations of PixelCNN \cite{fast-pixelCNN2,fast-pixelCNN1,anytime_fast-PixelCNN} can also be used to further speed up the auto-regressive generation procedure. In the current implementation of \detector, we adopt the fast pixelCNN from~\cite{fast-pixelCNN1}. 

%$m_{1} = 8$, $m_{2} = 16$, $m_{2} = 24$ and $m_4$ is the concatenation of the three views generated previously.Then, the final copy $G$ is created by copying its rows $r_{i1}$ to $r_{i2}$ from $G_{i}$. Assume an image is 32-by-32 in size and $G_{0} = x$. In our experiment, we set $(r_{01}, r_{02}) = (1,8)$, $(r_{11}, r_{12}) = (9,12)$, $(r_{21}, r_{22}) = (13,16)$, $(r_{31}, r_{32}) = (17,24)$, and $(r_{41}, r_{42}) = (25,32)$.

The view generation approach described above is static and does not vary from sample to sample. Whereas, the region of interest during view generation is the object location, and for each input sample it will be different. Currently, in static approach we generate three views with the expectation that object will be located in any one of them. On the other hand, if object location is known beforehand, in theory \detector~will only generate views that cover the object. However, problem with this approach is the white-box adversary may fool the algorithm that identifies region of interest. It is our future plan to explore and employ object localization approaches that are robust against white-box attacks. The localization approach need to be unsupervised for wider applicability and for \detector~we only need rough estimate of object location.

\subsection{Adversary Detection}
\label{sec:detection}
For adversarial detection, we have used four metrics to make our detector robust against different types of attacks. The metrics are discussed below. Our fundamental approach is to measure the inconsistency between an input image and its corresponding generated views. The intuition is that while the features learned by a classifier are a combination of both robust and non-robust features, the adversary can only tamper with the non-robust features to induce a misclassification of the target label~\cite{ilyas2019adversarial}. With the generative approach and the predicted label, \detector~ maps the robust features of the adversarial content (e.g., the object corresponding to the adversarial label) into generated views. Specifically, among the generated views, $G^*$ is more likely than others to capture the robust features of two distinct objects. Moreover, for an adversary example whose likelihood distribution $P(x')$ is lower than benign images, the generated pixels will be erroneous which means perturbation will be propagated in auto-regressive generative process. 

\vspace{1mm}
\noindent\textbf{Predictors.} 
\label{preds}The first metric measures the euclidean distance between the representation vectors of input image z and its generated view $G^*$, i.e:
\begin{equation}
\label{eq:d1}
    D_1(z)= || h(z) - h(G^*) ||_{2} 
\end{equation}
$D_1(z)$ is especially effective in detecting aggressive attacks that adds high perturbations. 

The second metric measures the distance in probability space with Kullback-Leiber(KL) divergence \cite{Joyce2011_KL} using output probability vector $f(.)$ of input image $z$ and its generated views $G_i$:
\begin{equation}
\label{eq:d2}
    D_2(z) = \sum_{i=1}^{3}\operatorname{KL}(f(z),f(G_i)) + \operatorname{KL}(f(z),f(G^*)) 
\end{equation}
$D_2(z)$ is highly effective when the added perturbation is small. 

The third metric is adopted from existing literature that only uses a single view i.e., the detector of PixelDefend (PD)~\cite{song2018pixeldefend}. It leverages the likelihood distribution of the input images to identify adversarial images. Since our view generation approach already leverages the likelihood distribution of images, the incorporation of this metric is straightforward:
\begin{equation}
\label{eq:d3}
    D_3(z) = P(z) 
\end{equation}

The fourth metric is adopted from literature, i.e., the I-defender (ID)~\cite{zheng2018robust}, which only uses the class conditional likelihood distribution of the classifier's representation layer $h(z)$ learned with the class conditional Gaussian mixture model (GMM):
\begin{equation}
\label{eq:d4}
    D_4(z) = P(h(z)|F(z)) 
\end{equation}
Since, it's obvious all four metrics are determined for particular image $z$, we will omit $z$ and use notation $D_i$.

\begin{figure}
\centering
    \includegraphics[width=0.98\columnwidth]{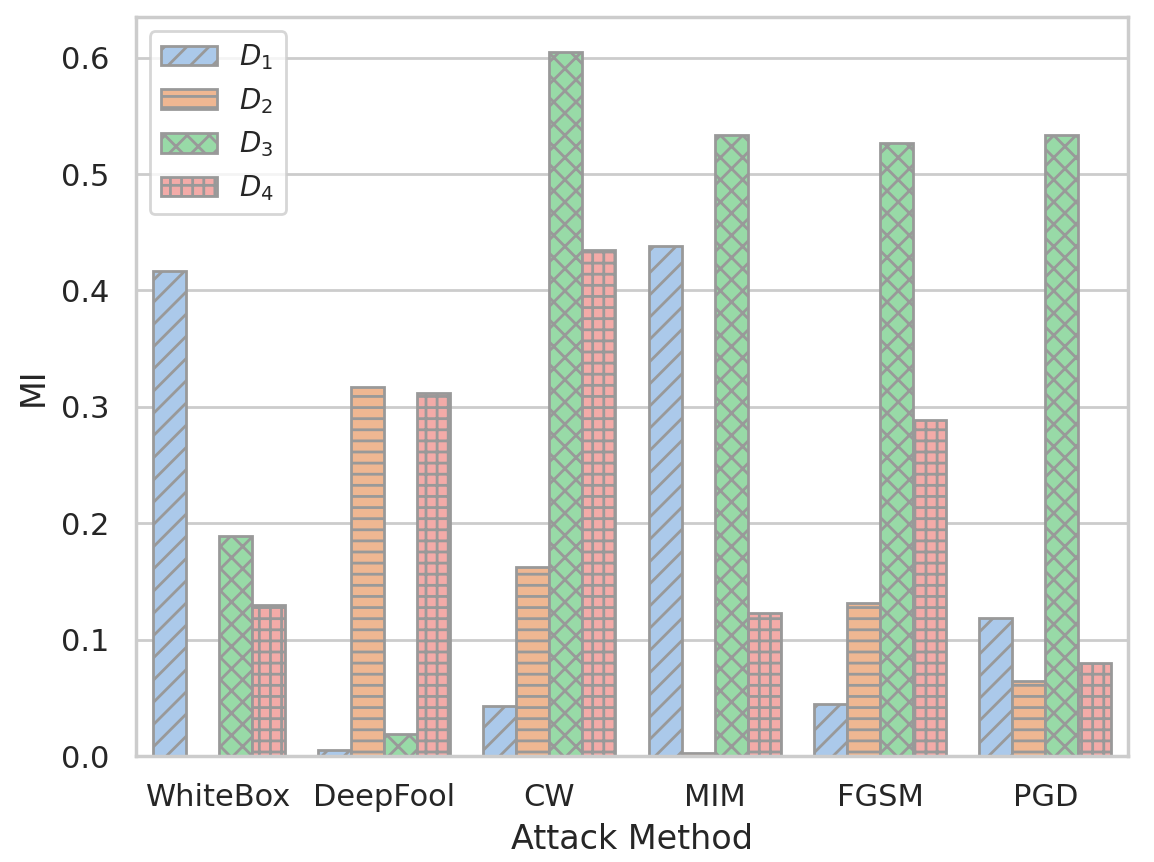}
\caption{Feature Importance for detection against various attacks. MI: mutual information. } \label{fig:MI}
\end{figure}

In previously proposed approaches \cite{Cohen_2020_CVPR,song2018pixeldefend,zheng2018robust}, a single metric was preferred to completely avoid training. However, it's difficult to guarantee general robustness due to varying nature of attacks \cite{Tramer_20_AdaptiveAttacks}. Based on our observation, different attacks have varying impacts on the metrics $D_1$ to $D_4$. % as shown in Figure \ref{fig:MI}. 
Roughly, we can divide the attacks into three categories: (i) the stealthier attacks with low perturbations, (ii) the aggressive attacks with high perturbations, and (iii) the defense-aware attacks (i.e., white-box attacks). We incorporated four metrics into our detection scheme that can counter all attacks effectively. To evaluate the importance of the above-mentioned predictors ($D_1$ - $D_4$) in detecting several representative attacks, we plot the mutual information (MI) between each predictor and detection label in Figure \ref{fig:MI}. As shown in the figure, the predictor $D3$ is effective against aggressive attacks like CW and iterative Linf-based attacks (MIM,PGD) but it is ineffective against DeepFool that is known to add small perturbations. Similarly, the metric $D_4$  is ineffective against iterative attacks (MIM, PGD). While each metric $D_3$ or $D_4$ is only effective against a certain type of attack and neither performs well against white-box attacks. However, newly proposed metrics $D_1$ and $D_2$ provide complementary information to $D_3$ or $D_4$ to boost the detection performance, and are also effective against white-box attacks (See Table~\ref{tab:Ab Study}). For instance, $D_1$ is very effective in detecting white-box and iterative MIM attacks. Whereas, $D_2$ is more effective in detecting small added perturbations, e.g., in DeepFool attack.  

\vspace{1mm}
\noindent\textbf{The Detector.} Finally, the \detector~detector utilizes all four predictors $d(z) = (D_1,D_2,D_3,D_4)$ in adversarial image detection. The objective of the detector design is to make it effective against different types of attacks, including potentially unknown attacks, with minimum training effort. We can observe from Figure \ref{fig:MI} that feature importance varies among different attacks and creates a covariate shift problem. Therefore, a supervised model trained with samples specific to one attack may be ineffective against other attacks.

%Here we discuss the possible approaches to counter this problem. The first approach is to determine optimum detection thresholds for all predictors simultaneously. However, this is simple and achievable with a single metric but gets complicated with each additional metric because they need to be tuned simultaneously to achieve certain benign detection rate. The other approach is to train a novelty detector like one-class SVM or isolation forest \cite{Liu2008_IF} by using only benign samples. However, the performance of such detectors is lower compared to a supervised detector that uses both benign and adversarial samples for training. 

In \detector, we adopt a hybrid of a \textit{supervised detector} and a \textit{novelty detector}. We observe certain patterns in the predictor feature importance with the representative attacks. As shown in Figure \ref{eq:d1}, the detection of aggressive attacks like CW and L-inf based (MIM, FGSM, PGD) rely on all four features to a certain degree with metric D3 being the dominant one. The other extreme is attacks that add small perturbations such as DeepFool, whose dominant features are different. Therefore, we adopt a supervised Random Forest (RF) \cite{Leo_01randomforests} classifier with $K$ trees and train it using adversarial samples only from DeepFool and PGD ($\epsilon = 4$). Meanwhile, a Novelty detector, Isolation Forest (IF) \cite{Liu2008_IF}, is trained using only the benign samples. The supervised detector identifies most of the (known) black-box attacks, while the novelty detector suffices for detecting white-box attacks, including potentially unknown attacks.

%The hybrid detector training and prediction procedure is: First create labelled training set $ X \in \{(d(x_1),y_1'),\cdot\cdot,(d(x_n),y_n')\}$ with input $d(x_i) = (D_1,D_2,D_3,D_4)$ and label $y_i' \in \{0,1\}$, to train supervised Random Forest (RF) \cite{Leo_01randomforests} Classifier with $K$ number of trees. The adversarial samples $y'=1$ in Training set $X$ were calculated using only PGD ($\epsilon$ = 4) and DeepFool attacks. 

In detection, the final outcome of the hybrid detector is ``benign'' ($D(z, F(z))=0$) if both supervised and novelty detector are confident to classify $z$ as benign, otherwise, the output label is ``adversarial'' ($D(z, F(z))=1$). 
\begin{equation}
    D(z,F(z))=\left\{\begin{matrix}
0 & P_{RF}(y'=0) + P_{IF}(y'=0) > \tau.\\ 
1 & \text{otherwise}\\
\end{matrix}\right.
\end{equation}
where $\tau$ is the fixed threshold and $y'$ is the output of RF and IF,
\begin{equation*}
\begin{split}
& P_{RF}(y'=0) = \frac{ \sum_{k=0}^{K} \mathbb{I}(t_k(d(z)) = 0) }{K}\\
& P_{IF}(y'=0) = 2^{\frac{-E(t(d(z)))}{c}}
\end{split}
\end{equation*}
$P_{RF}(y'=0)$ is the proportion of Trees $t_k$ in RF, classifies $d(z)$ as 'benign' and $\mathbb{I}$ is the indicator function. $P_{IF}(y' = 0)$ is the anomaly score of IF in which $t(d(z))$ is the path length of observation $d(z)$, $c$ is the average path length of unsuccessful search in a Binary Search Tree. We prove empirically that such detection scheme generalizes well across different kinds of attacks (See Table \ref{tab:greybox}).

%The last two criteria are adopted from the literature, i.e., the detector of PixelDefend (PD)~\cite{song2018pixeldefend} and the I-defender (ID)~\cite{zheng2018robust}. The first leverages the likelihood distribution of the input images to identify adversarial images, which is effective against aggressive attacks. The second uses the likelihood distribution of the classifier's penultimate layer $h(x)$ learned with the class conditional Gaussian mixture model (GMM), which is effective when the added perturbation is low. While each technique is only effective against a certain type of attack and neither performs well against white-box attacks, they provide complementary information to assist the three novel criteria we have designed for robust detection. 

%We found that ID , whereas PD is effective against . However, both these techniques are not effective for all types of attacks and also against white-box adversaries. That's where detection criteria based on our generated views, will give some range of protection against white-box attacks. Also, it will complement PD and ID to maintain a robust detection rate against all attacks.

%  (we compare their performance under different attacks in Table~\ref{tab:greybox})

%% file: experiment.tex
\section{Experiments}\label{sec:exp}

In this section, we present the adversarial example detection performance of \detector~against six different attacks 
on three datasets. We compare \detector~with two adversary detectors in the literature, and further analyze the performance of these approaches.

\subsection{Experiment Setup}
\vspace{1mm}
\noindent\textbf{Datasets.} We evaluate \detector~on three datasets: the German Traffic Sign Recognition Benchmark (GTSRB) \cite{stallkamp2012man}, the R-ImageNet (R-ImageNet) \cite{ILSVRC15}, and CIFAR10 \cite{cifar}. The dimension of images are fixed as 32 by 32. Adversarial sample defense for the whole ImageNet dataset is known to be a challenging problem, both due to the hardness of the learning problem itself, as well as the computational complexity. R-ImageNet is a subset of the original ImageNet dataset, which is used in other defense literature as well, e.g., \cite{tsipras2018robustness,ilyas2019adversarial}, to give similar complexity but with lesser number of samples/classes. 
On each data set, we use its default split of training, validation, and testing data. The numbers are presented in Table \ref{tab:dataset}. 

One important difference among the datasets is the level of complexity for the generation process. For instance, in the GTSRB dataset, object shapes and orientations remain the same across all the samples and there is little background. In CIFAR, object shapes and orientations change among samples with different backgrounds. In R-ImageNet, along with different object shapes, orientations, backgrounds, the added complexity is the similarity between different labels. From the classifier performance demonstrated in Table \ref{tab:dataset}, we can see that the classification accuracy of R-ImageNet is only at 74\%. Note that the classification accuracy of each victim classifier is listed here as a reference, it does not impact the detection performance of \detector. For a fair comparison with other approaches, it is important that the same datasets are used across all the experiments. We deliberately select a complex susbset to demonstrate the limits of \detector~and other adversarial sample detectors. 

%\vspace{5pt}

%image samples can have little background as well, we investigated \detector\ on CIFA10 dataset with no pre-processing.
\vspace{2mm}
\noindent\textbf{Metrics.} We adopt three metrics that are widely used in the literature: (i) the \textit{attack detection rate} (ADR) of an adversarial example detector is defined as the proportion of correctly detected adversarial samples out of all adversarial samples, i.e., the \textit{recall} of adversarial examples or the \textit{true positive rate} (TPR); (ii) the \textit{benign detection rate} (BDR) is the proportion of correctly labeled benign images out of all benign images, i.e., \textit{the true negative rate} (TNR); (iii) we also use the standard ROC curve and AUROC score to evaluate 
the true positive rate (TPR) with varying false positive rates (FPR=1-TNR). 
\begin{table}
  \caption{Datasets and performance of the base models}
  \label{tab:dataset}
  \begin{center}
  \begin{tabular}{|l||c|c|c|c|}
\hline
      & No. of & Train/Val/Test & Classifier & NLL \\
      &Classes&Samples&Accuracy& (bits)\\\hline
    GTSRB & 43 & 39,209/2,630/10,000 & 98\% & 1.7\\
    R-ImageNet & 16 & 40,517/400/1200 & 74\% & 4.60 \\
    CIFAR10 & 10  & 45,000/5000/10,000 & 95\% & 2.94\\
    \hline
  \end{tabular}
  \end{center}
\end{table}
\vspace{2mm}
\begin{table*}[!ht]
  \caption{Performance comparison: the Attack Detection Rate (evaluated with fixed TNR=95\%) and AUROC score of \detector~(ours), I-Defender \cite{zheng2018robust}, and PixelDefend \cite{song2018pixeldefend}. Both scores are scaled to [0, 100]. SR: Attack Success Rate. }
  \label{tab:greybox}
  \begin{center}
  \begin{tabular}{|c|c|ccc|ccc|c|ccc|ccc|c|ccc|ccc|}
  \hline
      & \multicolumn{7}{|c|}{GTSRB}& \multicolumn{7}{|c|}{CIFAR-10} & \multicolumn{7}{|c|}{R-ImageNet}  \\
      & & \multicolumn{3}{|c|}{\small Detection Rate}& \multicolumn{3}{|c|}{\small AUROC} & & \multicolumn{3}{|c|}{\small Detection Rate}& \multicolumn{3}{|c|}{\small AUROC} & & \multicolumn{3}{|c|}{\small Detection Rate}& \multicolumn{3}{|c|}{\small AUROC}  \\
    {\small Attack-$\epsilon$} & {\small SR} & {\small ID} & {\small PD} & {\small Ours} & {\small ID} & {\small PD} & {\small Ours} & {\small SR} & {\small ID} & {\small PD} & {\small Ours} & {\small ID} & {\small PD} & {\small Ours} & {\small SR} & {\small ID} & {\small PD} & {\small Ours} & {\small ID} & {\small PD} & {\small Ours} \\  
    \hline\hline
    % No Attack &  & -   &  /80/80& & /80/80 \\\hline
%    $L2$-Based & &  &   &  & & & & & & & & &  \\
    DeepFool& 94 & 68 & \color{red}{47} & \textbf{92} & 90 & 85 & \textbf{98} & 96 & 53 & \color{red}{14} & \textbf{58} & 90 & 59 & \textbf{92}& 100 & \color{red}{29} & \color{red}{8} & \color{red}{\textbf{42}} & 79 & 55 & \textbf{87}\\
    CW: 0.5    & 66 & 60 & 73 & \textbf{97} & 89 & 91 & \textbf{99} & 87 & 83 & 98 & \textbf{99} & 95 & \textbf{99} & \textbf{99}&98 & \color{red}{24} & \textbf{77} & 70 & 72 & \textbf{94} & 91\\ 
    \hline 
%    $L_\infty$-Based &  &  &   &  & & & & & & & & &  \\
    PGD-4   & - & - & - & - & - & - & - & 95 & \color{red}{23} & \textbf{98} & 91 & 73 & \textbf{98} & 97 & 76 & \color{red}{16} & \color{red}{15} & \color{red}{\textbf{42}} & 64 & 74 & \textbf{79} \\
    PGD-8   & 57 & \color{red}{32} & 84 & \textbf{86} & 82 & 96 & \textbf{98} & 96 & \color{red}{23} & \textbf{99} & \textbf{99} & 74 & \textbf{99} & 98& 91 & \color{red}{17} & \color{red}{46} & \textbf{64} &  65 & \textbf{89} & 87\\
    PGD-16  & 90 & \color{red}{38} & \textbf{99} & 88 & 84 & \textbf{99} & \textbf{99} & 96 & \color{red}{24} & \textbf{99} & \textbf{99} & 75 & \textbf{99} & 98& 92 & \color{red}{21} & \textbf{96} & 84 & 69 & \textbf{98} & 95 \\ \hline
    MIM-4   & - & - & - & - & - & - & - & 90 & \color{red}{28} & \textbf{98} & \textbf{98} & 76 & \textbf{98} & \textbf{99}& 77 & \color{red}{18} & \color{red}{19} & \textbf{51} & 65 & 75 & \textbf{79}\\
    MIM-8   & 52 & \color{red}{32} & 79 & \textbf{86} & 82 & 95 & \textbf{99} & 96 & \color{red}{45} & \textbf{99} & \textbf{99} & 85 & \textbf{99} & \textbf{99}& 91 & \color{red}{23} & \color{red}{34} & \textbf{71} & 71 & 84 & \textbf{88}\\
    MIM-16  & 87 & \color{red}{38} & \textbf{97} & 88 & 83 & \textbf{99} & \textbf{99} & 96 & 54 & \textbf{99} & \textbf{99} & 89 & \textbf{99} & \textbf{99}& 92 & \color{red}{23} & \color{red}{34} & \textbf{72} & 71 & 84 & \textbf{89}\\\hline
    FGSM-4  & - & - & - & - & - & - & - & 74 & 63 & 96 & \textbf{99} & 90 & \textbf{98} & \textbf{98}&55 & \color{red}{23} & \color{red}{40} & \color{red}{\textbf{48}} & 71 & \textbf{81} & 79 \\
    FGSM-8  & - & - & - & - & - & - & - & 82 & 85 & \textbf{99} & \textbf{99} & 95 & \textbf{99} & \textbf{99}&78 & \color{red}{29} & 73 & \textbf{75} & 73 & \textbf{94} & 92\\
    FGSM-16 & 48 & 61 & 88 & \textbf{96} & 90 & 97 & \textbf{99} & 86 & 96 & \textbf{99} & \textbf{99} & 97 & \textbf{99} & \textbf{99}&87 & \color{red}{41} & \textbf{95} & 89 & 80 & \textbf{98} & 96\\\hline\hline
    WB   & 89 & \color{red}{33} & \color{red}{38} & \textbf{86} & 80 & 85 & \textbf{97} & 90 & \color{red}{1} & \color{red}{13} & \textbf{71} & 31 & 70 & \textbf{91}& 96 & \color{red}{1} & \color{red}{3} & \color{red}{\textbf{45}} & 18 & 45 & \textbf{71}\\

    % \multicolumn{2}{|c|}{$L_{inf}$-Based}  &  &  & & & \\
    % MIM &16  &  /92/96& /79/99 & & /80/99 \\
    % FGSM &4  &  /55/75&  /64/86& & /69/83 \\
    % FGSM &8  & /78/84 &  /65/96 & &/79/94  \\
    % FGSM &16  & /87/87 & /72/99 & & /84/99  \\\hline
    % Whitebox & -  & /76/90 &/38/15  & & /58/50  \\
    
    \hline
  \end{tabular}
  \end{center}
\end{table*}
\noindent\textbf{Victim Model and Training. } We use the Wide Residual Network (w-ResNet) \cite{Resnet} as the target classifier, which the adversarial examples aim 
to fool. Its testing accuracy against benign samples is presented in Table~\ref{tab:dataset}. Meanwhile, two generative models are trained for \detector: (1) PxelCNN++ \cite{salimans2017pixelcnn++} is employed to learn class conditional input distribution $P(z|y)$ for view generation. Its performance is measured by the average natural log-likelihood (NLL) given as bits per dimension, which is also presented in Table~\ref{tab:dataset}. (2) A Gaussian Mixture Model (GMM) with eight components learns the class conditional distribution of the representation, i.e. $P(h(z)|y)$. Finally, we trained \detector~ using validation samples and reported detection performance on test samples.

%\vspace{5pt}
\vspace{2mm}
\noindent\textbf{The Failed attacks.} When an adversarially modified image fails to trigger misclassification, it is a failed attack. For instance, an adversarial example may fail if the target model is robust or the added perturbation is too low. As shown in Table \ref{tab:greybox}, the success rates (SR) of adversarial attacks range from 100\% (DeepFool on R-ImageNet) to 48\% (FGSM with $\epsilon=16$ on GTSRB) or even lower (FGSM-4 and FGSM-8 on GTSRB, PGD-4 on GTSRB, etc). In this paper, we do not consider failed attacks in calculating the detection rates, which is a common practice in the literature.

\subsection{Experiment Design}
To evaluate \detector\ we generated adversarial examples based on two threat models presented in Section \ref{Attack_Model}. %Also, we compare our results with PixelDefend (PD) \cite{song2018pixeldefend} and I-defender (ID) \cite{zheng2018robust}.

\vspace{2mm}
\noindent\textbf{The Black-box Attacks: } The black box attack methods involved in our experiments are explained in Appendix \ref{sec:attacks}. We have used cleverhans implementation to generate these attacks \cite{papernot2018cleverhans}. For each attack method, all adversarial examples were limited by the maximum allowed perturbation i.e. $\left \| x-x' \right \|_p \leq \epsilon$, where $\epsilon \in [0,255]$. For attack methods whose distance metric is $L_\infty$, e.g., FGSM, PGD, and MIM, the added perturbation is controlled by $\epsilon$. On the other hand, among the attack methods that use $L_2$ metric, DeepFool ensures to add the smallest perturbation by design, while perturbations in C\&W could be controlled by the confidence parameter [0.1,0.9]. In our experiments, adversarial examples are generated with the default settings in DeepFool. C\&W attacks are generated with a confidence parameter of 0.5. Three perturbation levels $\epsilon=4,8,16$ are used for PGD, MIM, and FGSM.  

\vspace{2mm}
\noindent\textbf{The White-Box Attack:}  In order to generate a white-box attack against \detector, we design a method based on the adaptive techniques described in \cite{zheng2018robust,carlini2017adversarial,Tramer_20_AdaptiveAttacks}. \detector\ uses generated samples to perform detection based on predictors given in Eq. \ref{eq:d1} to \ref{eq:d4}. For white-box attack, first two predictors given in Eq. \ref{eq:d1}, \ref{eq:d2} can not be controlled since there is no direct access to the generated views. However, it can be ensured that $P(x'|F(x'))$ stay high like clean examples. Similarly, it can be ensured $P(h(x')|F(x'))$ remains high i.e. closer to benign samples. We use Iterative $L_{\infty}$ attack with perturbation level $\epsilon = 8$ to maximize the following function for $x$:
\begin{multline}
    \underset{x}{\operatorname{argmax}}[J(\theta,x,y_t) + \alpha\prod_{k \neq t}\operatorname{Reject}_1(x,y_k)\times \underset{k\neq t}{max}\log P(x|y_k) \\
    + \beta\prod_{k \neq t}\operatorname{Reject}_2(x,y_k)\times \underset{k\neq t}{max}\log P(h(x)|y_k)]
\end{multline}
where the first term $J(.)$ denotes the cross-entropy loss of classifying $x$ into its true label $y_t$ by a classifier with parameter $\theta$. This term encourages finding a perturbation that leads to misclassification. The second term penalizes $x$ if its likelihood score from any other classes is lower than the score of the clean sample from the true label. This term encourages $x$ to attain a high likelihood probability with the false label. Similarly, the third term encourages $x$ to attain a high likelihood distribution of the representation vector with the false label. Other possible way to generate white-box attack is using optimization based method \cite{carlini2017towards}. We have observed that \detector~is effective against both iterative and optimization based black-box attacks. We expect that if it also works for iterative white-box attack, it will work for optimization based white-box attack as well, since objective remain same in both methods.

\subsection{Experimental Results}\label{sec:exp_results}

\textcolor{black}{In this section, we report the experimental 
results of \detector~against the six attacks 
{on} three datasets, and 
compare them with I-Defender (ID) \cite{zheng2018robust} and  PixelDefend (PD) \cite{song2018pixeldefend}.}

\vspace{2mm}
\noindent\textbf{Detection Rate and AUROC.} We first evaluate the adversarial example detection rate (ADR) of all three approaches at a fixed TNR of 0.95. The experimental results are reported in Table~\ref{tab:greybox}. We also evaluate the AUROC score in each attack and present the results in the same table. We do not evaluate PGD and MIM with $\epsilon=4$, or FGSM with $\epsilon=4, 8$ on GTSRB since their attack success rates are too low (<<40\%) to be practical. 

The best performance for each experiment is shown in bold. \detector~performs the best (including ties) in 49 out of 64 experiments. \detector' overall average ADR across all experiments is 80.7\%, while the overall average ADR of ID is 37.7\% and the average ADR of PD is 67.4\%. The average AUROC score of \detector~is 0.934, while the average AUROC scores of ID and PD are 0.765 and 0.886, respectively.

We highlight the severe detection failures in red in Table~\ref{tab:greybox}, as they are 
the most destructive cases in practice where detection rates drop below 50\%. In particular, the two SOTA detectors can only identify 1\% to 3\% of the white-box attacks on the R-ImageNet 
data set and 1\% to 13\% on the 
CIFAR-10 data set. Comparatively, 
\detector~manages to identify 45\% on R-ImageNet and 
71\% on CIFAR-10, a substantial improvement 
over the literature. Besides, the two 
prior detectors have 23 and 12 severe 
failures respectively distributed in all three datasets, while \detector~only has 4 in R-ImageNet dataset, among which it still 
achieves the best performance. 
This demonstrate the outstanding 
robustness of \detector~in worst-case scenarios.

\vspace{2mm}
\noindent\textbf{Overall System Performance.} The overall accuracy of a classifier+detector system (Figure \ref{fig:overview}) is defined as the proportion of correctly classified samples and correctly identified adversarial samples out of all inputs. For classifier+detector, there exist a trade off between drop in performance for benign examples and gain in performance for adversarial examples. We have set \detector~TNR = 95\%, to allow 5\% performance drop for benign samples. 

%We consider an extreme case where all the input samples are adversarial. 
Without a detector for adversarial examples, the classification accuracy is the portion of failed attacks that are correctly classified over all attacks. With the existence of \detector, a significant portion of the attack samples are detected. Table \ref{tab:overall} presents the classification accuracy of the victim DNN against adversarial samples, and the overall classification+detection accuracy of the DNN+\detector~against the same input. 
%However, the gain in case of adversarial examples is much higher as presented in Table \ref{tab:overall}. 
Across all the experiments, \detector~raises the system accuracy from 14.0\% to 81.3\%. 

\begin{figure*}
\centering
\begin{tabular}{ccc}
    \includegraphics[trim=66 00 66 40,clip,width=0.66\columnwidth]{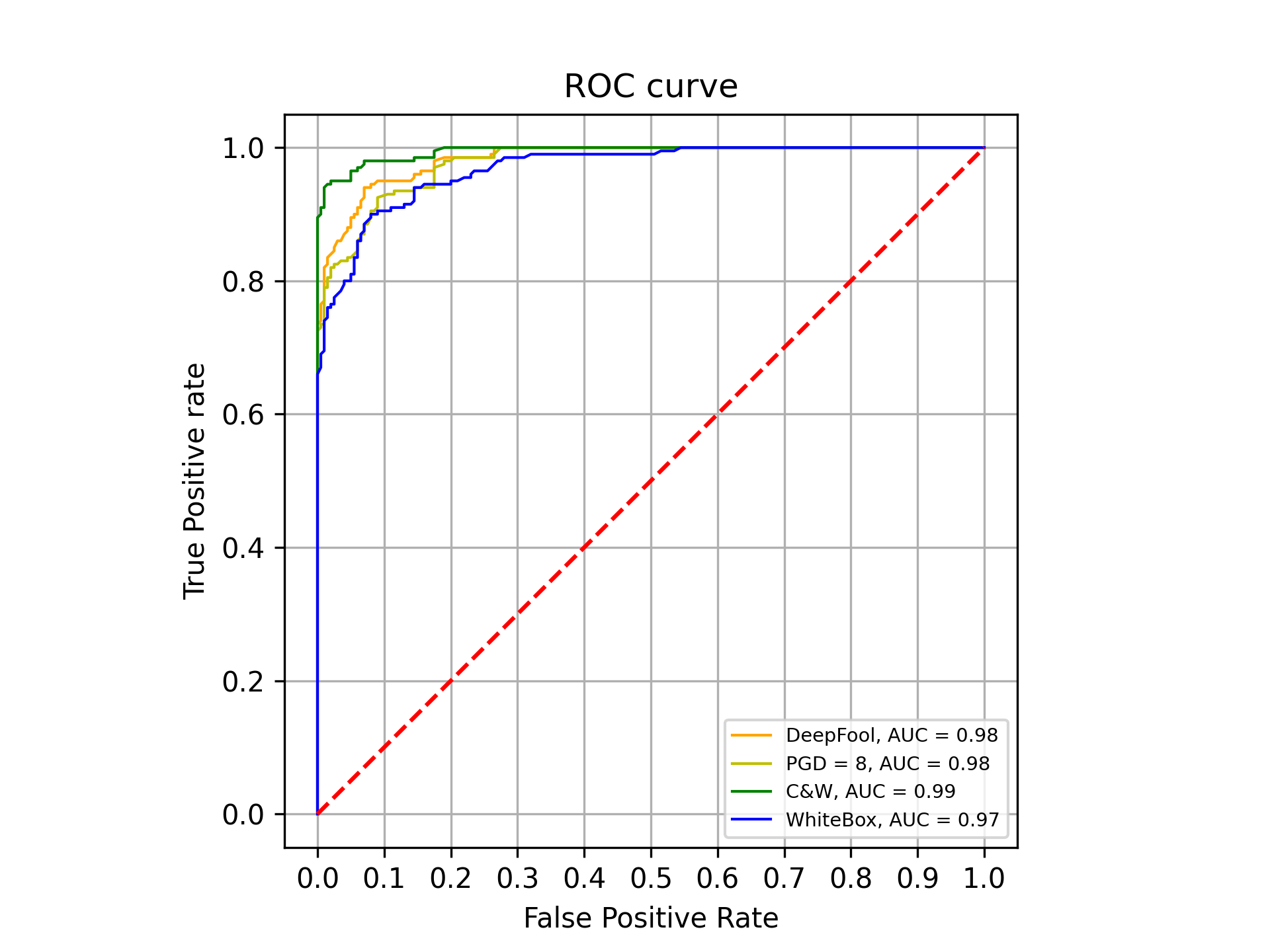} & \includegraphics[trim=66 00 66 40,clip,width=0.66\columnwidth]{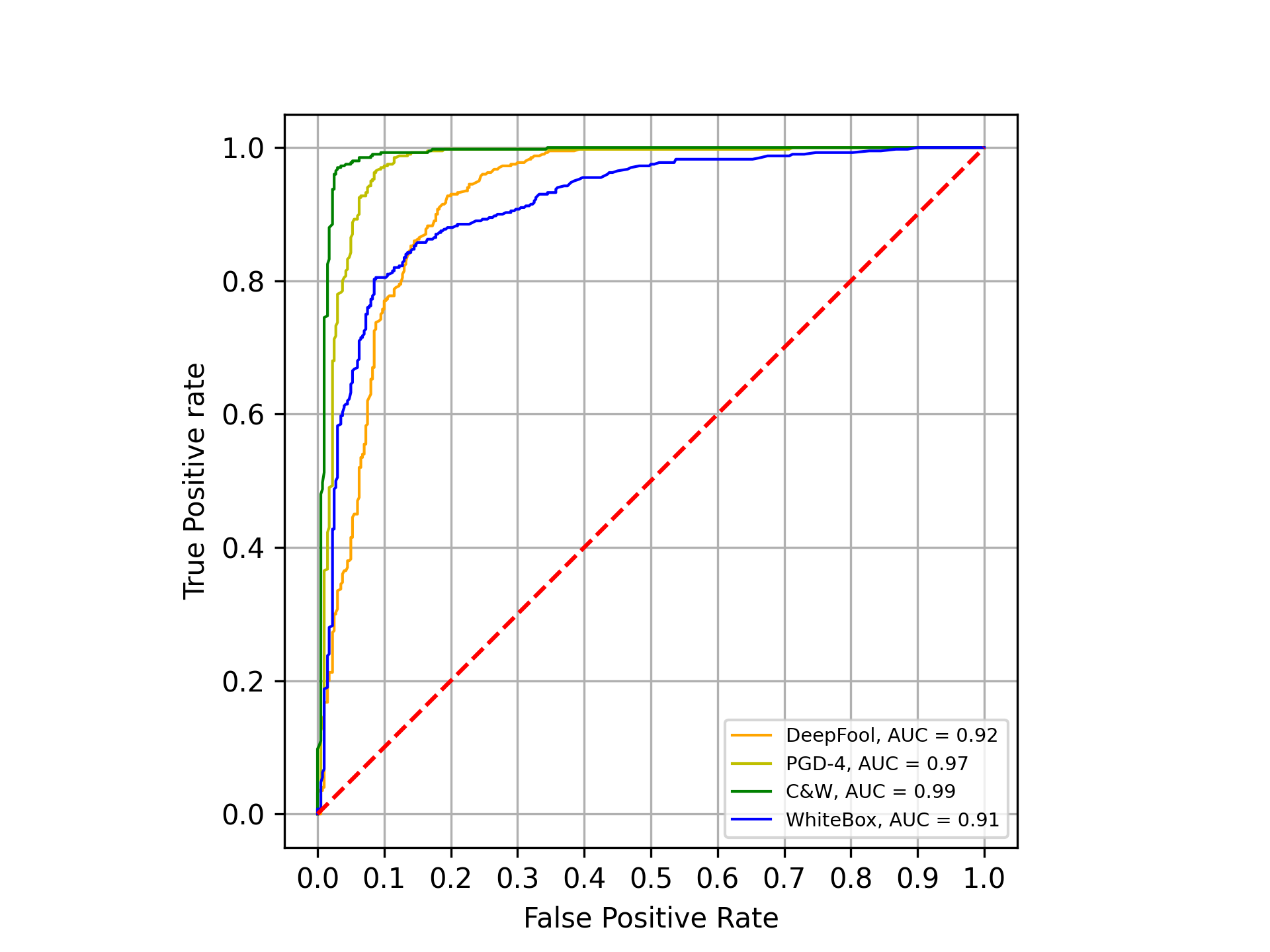} &  \includegraphics[trim=66 00 66 40,clip,width=0.66\columnwidth]{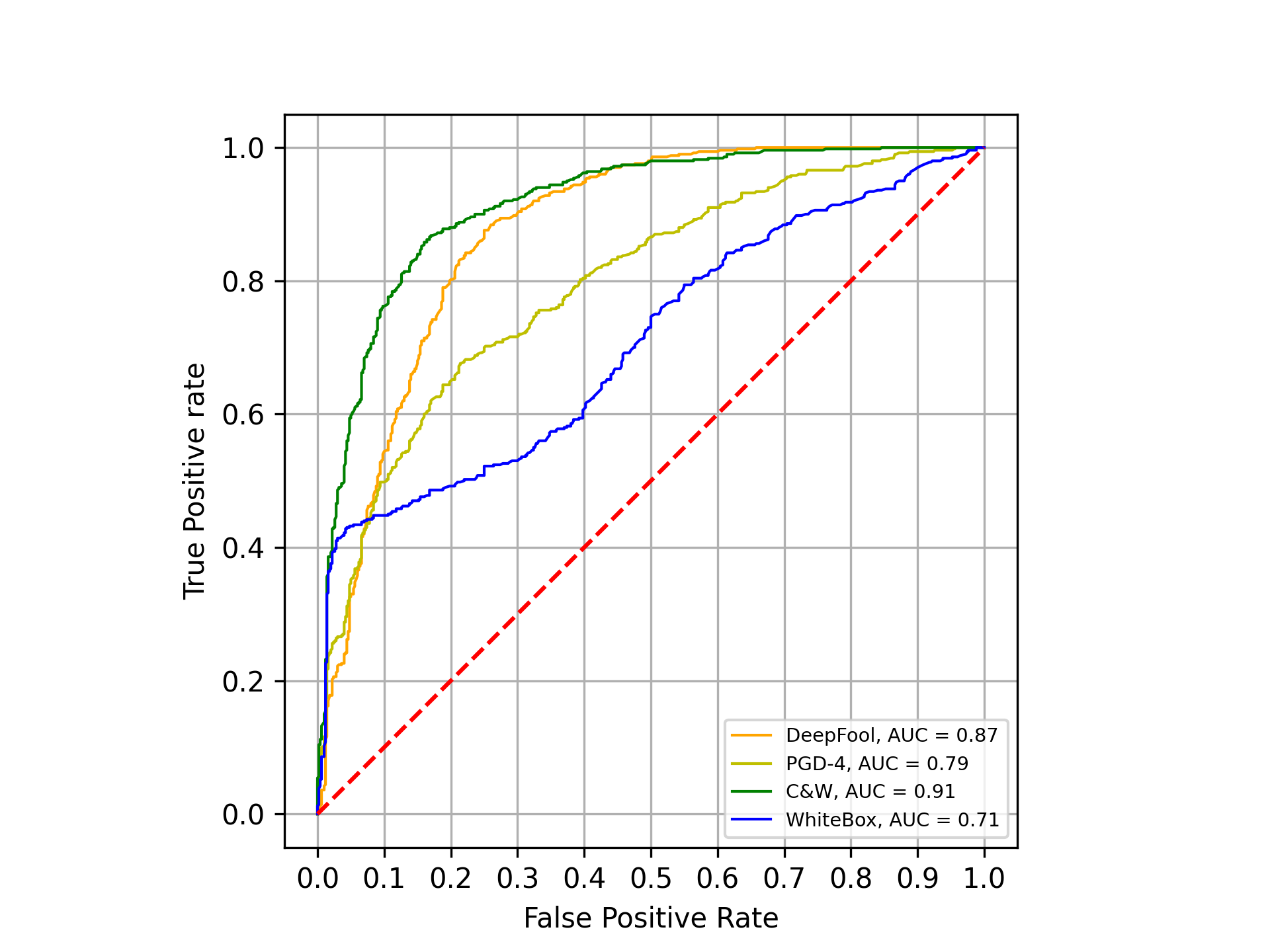} \\
    (a) & (b) & (c)
\end{tabular}
\caption{ROC Curves against white box and small perturbation attacks: (a) GTSRB, (b) CIFAR-10, (c) R-ImageNet. } \label{fig:roc}
%\vspace{-2mm}
\end{figure*}

\vspace{2mm}
\noindent\textbf{ROC.} Finally, we evaluate \detector~against the attacks that are most difficult to detect: (1) DeepFool; (2) PGD with $\epsilon$=4; (3) C\&W; and (4) the white-box attack. To better demonstrate \detector~detection accuracy over varying thresholds, we present the ROC curves in Figure~\ref{fig:roc}. We can clearly see that \detector~provides solid performance even against the strongest or most stealthy attacks in GTSRB and CIFAR-10 datasets. The performance in R-ImageNet is not as impressive but it's quite good given the complexity of the dataset. 

\vspace{2mm}
\noindent\textbf{Ablation Study.} In Table \ref{tab:Ab Study}, we present detection performance for the CIFAR10 dataset by using different combinations of the predictors presented in Section \ref{sec:detection}. Although a single predictor has the advantage that it is easier to adjust the threshold value, we observe from the results in Table~\ref{tab:Ab Study} that good generalization can not be guaranteed using a single predictor--the AUC scores suffer by a wide margin compared to the best performances, which are achieved using a combination of predictors. For instance, $D_1$'s performance against DeepFool is $54\%$, which is much lower than the best performance, $92\%$,  which is achieved with $D_1$, $D_2$ and $D_4$. 

The high performance across all attacks is guaranteed when all four predictors are used. However, we note that there are also a couple of three-predictor configurations, such as $(D_1, D_2, D_3)$ and $(D_1, D_3, D_4)$, whose performance stays consistently high (in the close vicinity of the best performance) among all attacks. In this case, it is possible to utilize the combination of $D_1$, $D_2$ and $D_3$ for efficiency, as $D_4$ requires training of another generative model.

%We conclude from the extensive experiments that \detector~significantly outperforms two representative adversarial detectors in both detection accuracy and robustness. While \detector~does not rely on any prior knowledge nor makes any assumption about the adversarial attack method, it demonstrates consistently high performance against a wide spectrum of attacks. 

%For R-ImageNet dataset, we used the detection rule mentioned in Eq. \ref{detector2} . For clean samples detection rate set to $95\%$, the adversarial detection performance is given in Table. 

%Since, generative process for clean samples will not be perfect for this dataset, due to which clean detection drops to $85\%$ dow The evaluation results are presented in Table. 

%The proposed detector perform really well for GTSRB dataset against all type of attacks. As, each traffic sign will have similar appearance (shape, color and orientation) among all samples. Hence, the quality of generated views will be quite high. For more complex datasets like ImageNet and CIFAR10 where each object appearance changes from sample to sample, the proposed scheme still performs reasonably well. 

\vspace{2mm}
\noindent\textbf{Detection Throughput.}
On a single node of Nvidia TITAN Xp GPU, \detector~ takes approx. 107 seconds to process 100 images. The major bottleneck in throughput is the autoregressive generative process. which generates the three base views, i.e. $G_1 - G_3$. Since all three views are independent, they can be generated in parallel if more computational nodes are available, which will eventually speed up the computation by three folds approximately. %With three nodes of k80 GPU, \detector~ detection rate is approx. 70s for 100 images. 
%\vspace{-2mm}

\begin{table}
  \caption{System accuracy against adversarial inputs.}
  \label{tab:overall}
   DNN: classification accuracy of the the victim DNN; \\ +\detector: accuracy of DNN+\detector~ (adversarial labels are rejected). \vspace{1mm}
  \begin{center}
  \begin{tabular}{|c|cc|cc|cc|}
\hline
    & \multicolumn{2}{|c|}{GTSRB} & \multicolumn{2}{|c|}{CIFAR-10} & \multicolumn{2}{|c|}{R-ImageNet}\\

    Attack-$\epsilon$ & {\small DNN} & {\small +Argos} & {\small DNN} & {\small +Argos} & {\small DNN} & {\small +Argos}\\
  
    \hline\hline
%No Attack	&	98	&	93.1	&	95	&	90.2	&	74	&	70.3	\\\hline 
DeepFool	&	5.9	&	92.1	&	3.8	&	59.3	&	0.0	&	42.0	\\
CW: 0.5	&	33.3	&	95.7	&	12.4	&	97.9	&	1.5	&	70.0	\\\hline
PGD-4	&		&		&	4.8	&	91.2	&	17.8	&	48.8	\\
PGD-8	&	42.1	&	89.1	&	3.8	&	98.7	&	6.7	&	64.6	\\
PGD-16	&	9.8	&	88.5	&	3.8	&	98.7	&	5.9	&	82.9	\\\hline
MIM-4	&		&		&	9.5	&	97.2	&	17.0	&	55.4	\\
MIM-8	&	47.0	&	89.4	&	3.8	&	98.7	&	6.7	&	71.0	\\
MIM-16	&	12.7	&	88.7	&	3.8	&	98.7	&	5.9	&	71.9	\\\hline
FGSM4	&		&		&	24.7	&	96.7	&	33.3	&	58.0	\\
FGSM8	&		&		&	17.1	&	97.4	&	16.3	&	74.0	\\
FGSM16	&	51.0	&	94.5	&	13.3	&	97.8	&	9.6	&	86.6	\\\hline\hline
WB	&	10.8	&	86.8	&	9.5	&	72.9	&	3.0	&	46.0	\\

    \hline
  \end{tabular}
  \end{center}
\end{table}

\begin{table}
  \caption{AUC(\%) Detection performance of different predictors on CIFAR-10}
  \label{tab:Ab Study}\vspace{-1mm}
    %\vspace{1mm}
  \begin{center}
  \begin{tabular}{|cccc|cccc|}
\hline
    \multicolumn{4}{|c|}{Predictors}&\multicolumn{4}{|c|}{Attacks}\\
    \hline\hline
    $D_1$&$D_2$&$D_3$&$D_4$&DF&PGD-$\epsilon_4$&CW&WB\\ \hline
    $\checkmark$&&&&54&77&67&\textbf{93}\\ \hline
    &$\checkmark$&&&84&64&73&53 \\\hline
    &&$\checkmark$&&59&98&\textbf{99}&70 \\ \hline
    &&&$\checkmark$&90&73&95&31 \\ \hline
    $\checkmark$&$\checkmark$&&&87&73&76&87\\ \hline
    $\checkmark$&&$\checkmark$&&55&\textbf{99}&\textbf{99}&87\\ \hline
    $\checkmark$&&&$\checkmark$&86&78&95&78\\ \hline
    &$\checkmark$&$\checkmark$&&85&96&\textbf{99}&58\\ \hline
    &$\checkmark$&&$\checkmark$&91&74&94&42\\ \hline
    &&$\checkmark$&$\checkmark$&84&97&90&40\\ \hline
    $\checkmark$&$\checkmark$&$\checkmark$&&87&97&\textbf{99}&85\\ \hline
    $\checkmark$&$\checkmark$&&$\checkmark$&\textbf{92}&79&93&81\\ \hline
    $\checkmark$&&$\checkmark$&$\checkmark$&85&98&\textbf{99}&80\\ \hline
    &$\checkmark$&$\checkmark$&$\checkmark$&91&96&91&51\\ \hline
    $\checkmark$&$\checkmark$&$\checkmark$&$\checkmark$&\textbf{92}&97&\textbf{99}&91\\ \hline
  \end{tabular}
  \end{center}
\end{table}

\subsection{Analysis of Experiment Results}\label{sec:expanalysis}
\vspace{1mm}
\noindent\textbf{Attack Performance.} When we compare the attack performance across three datasets, the attack success rates are generally affected by two factors: the complexity of the images and the distance between images from different categories. Attacks on GTSRB are the least successful since images in this dataset are relatively simple--each image has a clear traffic sign in the center and very little background. Moreover, images in each class are quite distinctive from images in other classes. As a result, the image classifier achieves 98\% accuracy on this dataset while evasion attacks are difficult. In particular, PGD, MIM and FGSM attacks with low perturbation (e.g., $\epsilon=4$) achieve very low ASRs. 

\vspace{2mm}
\noindent\textbf{Detection Performance.} Compared with SOTA adversarial image detectors, \detector~generates higher adversarial detection performance against almost all the attacks and datasets. In particular, for datasets like GTSRB and CIFAR-10 whose classification accuracy is higher,  \detector~demonstrates superior performance (~90\% ADR). The classification task is simpler when different classes are distinct from each other, which means the two ``souls'' that \detector~exploits for detection are far apart. The results with CIFAR also demonstrate that \detector~is not limited by the complexity of images like different object positions, shapes, orientations, and backgrounds. The only limitation comes when the two ``souls'' of the benign and adversarial labels are not quite different, e.g., an adversarial ``truck'' image mislabeled as ``automobile''. This is the reason why \detector~perform comparatively lower in R-ImageNet dataset, as even the classifier is struggling to differentiate the objects for benign images.

\vspace{2mm}
\noindent\textbf{Generative Models.} The class conditional distribution of representation vector $P(h(x)|y)$ is used to identify adversarial samples in~\cite{lee2018simple} and I-Defender \cite{zheng2018robust}, since the representation of each sample in one class resides in close vicinity to form a group, but the representation of the adversarial samples may not belong to any group. This approach is especially effective against the DeepFool attack that add very small perturbations to push the representation of the adversarial samples towards the target class to trigger misclassification, while the perturbation is not strong enough to push the representation to the close vicinity of the benign samples of the target group. However, this approach cannot guarantee similar performance with more aggressive and white-box attacks, which attempt to push the the adversarial samples to the target group.  

On the other hand, PixelDefend (PD)~\cite{song2018pixeldefend} uses the distribution of input samples $P(x)$ to identify adversarial samples. Like other approaches~\cite{metzen2017detecting,samangouei2018defensegan} that use input images for detection, PD is ineffective when the added perturbation is small. Overall, the generative models used in detection are ineffective against white-box attacks since the adversary can ensure high likelihood distribution for the adversarial samples. The major difference in our approach is that we use generative models for view generation instead of likelihood estimation to incorporate predicted label information effectively.

Last, the R-ImageNet is the most complex dataset in our experiments. The complexity in the dataset also introduces additional difficulties in the generative model. As a result, all the generative-model-based detectors perform worse on R-ImageNet data.

%% file: discussions.tex
\section{Discussions}\label{sec:dis}
\vspace{1mm}
\noindent\textbf{The Interpretability of \detector}. In \detector, the predictors $D_1$ to $D_4$ are used to highlight inconsistencies between the seed image and the views (Section \ref{sec:detection}). The detection properties of these hand-picked predictors are beneficial for three reasons as compared to using a DNN as a detector: (1) The proposed detector in Section \ref{sec:detection} is more interpretable than a neural network-based model. The response of the adversarial examples to the predictors may give us an insightful understanding of the attacks. (2) Each of the predictors' outputs can be interpreted as a distance metric which is small for benign samples but increases as the adversarial perturbations increase. Thus, a one-class classifier can be used in conjunction with these predictors for adversarial detection. On the other hand, training a supervised classifier would only require adversarial examples that lie close to decision boundary (i.e. attack samples that add the smallest perturbation) while the classifier would work well as perturbation increases. While we can certainly employ a multi-view classification model for this task, a significant amount of captured or synthetically generated adversarial samples will be needed in training which still cannot guarantee similar performance against unknown attacks. And (3) we intentionally choose a simple detector to highlight the effectiveness of the generation-and-detection approach that amplifies image-label discrepancies and the effectiveness of the novel concept of adversarial detection using multi-view inconsistency. Experiment results have shown that a relatively simple detector is capable of effectively detecting adversarial examples. 

% \vspace{5pt}
% \noindent\textbf{} 
%The main objective of \detector~is to ensure that detection performance remains consistently high even for the most stealthy and previously unknown attacks. 
\vspace{1mm}
\noindent\textbf{Generalization and the White-box Attack Revisited. } 
As we have discussed in Section \ref{sec:solutionoverview}, \detector~does not make any assumption of the attack method, except the fact that the visual content of the testing image and the label it receives are inconsistent. In our experiments, when \detector~is trained using only the adversarial examples from DeepFool and PGD-4, we see in Table \ref{tab:greybox} that it generalizes well for all attacks including white-box attacks, which implies that: (1) although the supervised detector is only trained with two attacks, it is capable of detecting a wider range of attacks; and (2) the \textit{novelty detector}, as a part of \detector, plays a significant role in identifying and mitigating \textit{unknown} attacks.

When the defensive mechanism against evasion attacks makes any assumption on the attack method, a white-box attack could always be designed to challenge the assumption and thus produce highly successful attacks. To defend against white-box attacks, the only reliable assumption one could make is that the attacker's goal is to trigger misclassification. Therefore, we incorporate the classified label from the DNN in the design of the detection algorithm. Intuitively, we can simply learn the conditional likelihood distribution $p(x|y)$ and expect $p(x|\text{Correct Label}) > p(x|\text{Adversarial Label})$. However, this would not perform well due to the high dimensionality and complexity of image data, as shown in \cite{van2016pixel}. In \detector, we employ the autoregressive generative models to construct views and sample the conditional probability distributions from there. 
%As we have discussed, \detector~is expected to be more robust against the white-box attacks since it makes no assumptions on the attack method. While the adversary could still attack two factors in the decision criteria, as we have demonstrated in the proposed white-box attack in Section \ref{sec:exp}, she cannot deviate from her goal, which is to flip the label. In response, \detector~heavily relies on the (possibly flipped) label in its view generation and decision process. \detector~has demonstrated potential in combating white-box attacks, as shown in Table \ref{tab:greybox} and Figure \ref{fig:roc}. %It is our ongoing research to further improve \detector~against unknown and white-box attacks. 

%As shown in Figure \ref{fig:gen_images}, first two columns of generated samples contain classified label objects. So, it's one effective way to incorporate classified label information that can be effective for white box attacks. However, they are also helpful in detecting black box attacks as well.

%For Black box attacks, there are number of ways to amplify perturbation like image resizing, interpolation etc. However, as shown in Figure the likelihood probability of adversarial examples is low as they don't exist naturally[pixel defend]. To make it consistent with our detection approach, we will again use autoregressive generation process. The idea is when likelihood probability of seed pixels is low which should be in case of adversarial samples, the generated pixels will be noisy and our objective to create dissimalrity with candidate sample will be achieved.    
\begin{table}
  \caption{Detection accuracy (AUC(\%)) of misclassified benign vs. adversarial samples.}
  \label{tab:missclassify detection}\vspace{-3mm}
  \begin{center}
  \begin{tabular}{l||c|c|c|c}
%\hline
   & \multicolumn{2}{c|}{Misclassified Benign} & \multicolumn{2}{c}{Adversarial}\\
    Detector & CIFAR & R-ImageNet & CIFAR & R-ImageNet \\\hline
%    $D_1$ + $D_2$ & 81.2 & 72.4 & 82.6 &72.7\\
    \detector & 83.8 & 74.2 & 97.1 & 85.1 \\
 %   \hline
  \end{tabular}%\vspace{-3mm}
  \end{center}
\end{table}
\vspace{1mm}\noindent\textbf{Detection of Misclassified Benign Samples. } By design, the predictors $D_1$ and $D_2$ in \detector~exploit the misclassified label in order to detect the adversarial samples. Using the same principle, the predictors are supposed to be effective in identifying misclassified benign samples (a.k.a., naturally misclassified samples). As shown in Table \ref{tab:missclassify detection}, \detector~is able to detect approximately 75\% to 83\% of naturally misclassified samples. This phenomenon of detecting misclassified benign samples by adversarial detector is common in other detection approaches as well \cite{zheng2018robust,Vacanti_detector_2020}. In safety-critical applications, it is desired if the adversarial detector works in a semi-autonomous fashion and reports all types of misbehaviors \cite{metzen2017detecting}. 

\begin{figure}
\centering
    \includegraphics[width=\columnwidth]{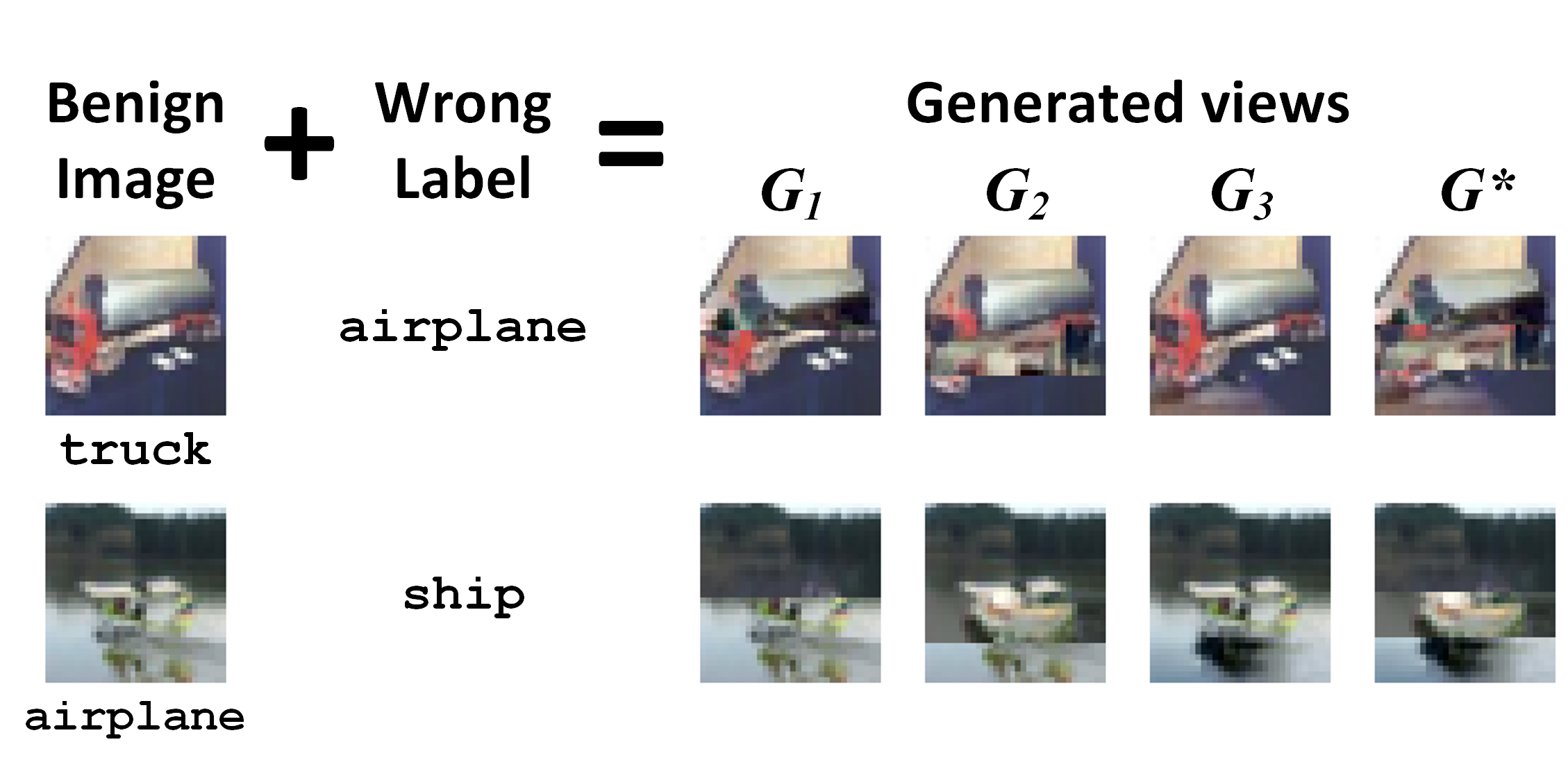}
\caption{Misclassified benign samples (CIFAR): (1) truck image misclassified as airplane; %(2) truck image misclassified as automobile; 
(2) airplane image misclassified as ship. \detector~is able to detect (2) as misclassified. } \label{fig:benign}
\end{figure}

In \detector, we only make one assumption of the input image and label: there exists an inherent discrepancy between the visual content of the image and the label, so that \detector~could further amplify and assess such discrepancy for adversarial image detection. This assumption is guaranteed to be true for all successful attack images. Meanwhile, this assumption is also true for a portion of the naturally misclassified images. However, the fundamental difference between naturally misclassified samples and adversarial samples is that the naturally misclassified samples are more likely to carry visual features akin to the wrong label, so that the image-label discrepancies are low. Figure \ref{fig:benign} demonstrates two misclassified benign samples from the CIFAR dataset. The tanker truck in the first image has a metal tank that is visually similar to a wing of an \texttt{airplane} (the misclassified label). Therefore, the generative model was able to produce reasonably consistent ``airplane'' views from the source image. As a result, \detector~was unable to detect this misclassified sample. 
 
%With such images, the view generation approach 
Finally, we reiterate that the essential objectives of the adversarial attacks are (1) to add minimum perturbation to $x$ so that it still looks like its original label $y$, and (2) to trick the classifier to label the image as a significantly different label $y'$. Both objectives work in the favor of \detector. As shown in Table \ref{tab:missclassify detection}, \detector~is significantly more effective in detecting adversarially misclassified samples. 

%However, if it is desirable to further identify adversarial interference in the system, the achieved objective with misclassification can be analyzed further, since misclassified benign samples are not expected to effect the system integrity in the same manner as adversarial samples. We hope to explore this aspect of adversarial detection as part of future research.
\vspace{1mm}\noindent\textbf{Limitations.} \detector~does not perform well when the visual content of the source class $y$ and the misclassified class $y'$ are not too distinct to each other, such that even a conventional classifier struggles to differentiate between the two classes of benign samples. For instance, a long-hair cat might look visually similar to a dog when observed from a particular angle. When the attacker uses such a cat image to force a ``dog'' label, \detector~is less likely to detect the attack. However, for security-critical applications, where miss-classification even for benign samples is undesirable, different classes are usually quite distinct like traffic signboards. Thus, we expect \detector~will perform well in security-critical applications. 

Last, as shown in Table \ref{tab:greybox}, it is also evident that \detector'~performance is relative to the complexity of the dataset. For a dataset whose benign classification accuracy is high, such as GTSRB and CIFAR10,  \detector' detection AUC score is consistently higher than $90\%$. Meanwhile, for a complex dataset, such as R-ImageNet, whose classification accuracy is approximately $74\%$, the \detector~guarantees AUC score above $70\%$ ($>80\%$ against most attacks), which is acceptable given that the misclassification rate of the benign classifier is already high. We argue that the problem of adversarial examples usually arises in scenarios where misclassified examples can significantly damage the system integrity, i.e., when the natural classification rates with benign samples are much higher.

\vspace{1mm}\noindent\textbf{Detection Performance for larger Images.} In the literature of adversarial example detection, it is standard practice to do proof of concept with several popular image datasets, such as MNIST, F-MNIST, GTSRB, CIFAR, Tiny-ImageNet. To perform a fair comparison with the literature, especially I-Defender \cite{zheng2018robust}, and PixelDefend \cite{song2018pixeldefend}, we used the same experimental setup that employs small images (i.e., 32$\times$32) in evaluation.  Moreover, we would like to ask how \detector~would perform for larger images. Hence, we further evaluate \detector~with 64$\times$64 images from the Restricted-ImageNet dataset. The classification accuracy and pixel-cnn's natural log-likelihood (NLL) achieved for this dataset are 79\% and 3.97 respectively, both of which are slightly improved over 32$\times$32 images (74\% classification accuracy and 4.60 NLL, as reported in Table \ref{tab:dataset}).
Comparing with experiment results on smaller images, the detection performance of \detector~has improved for iterative attacks, such as PGD, MIM and WB, as shown in Table \ref{tab:detect_large}. However, for optimization based attacks like CW and single step FGSM, the performance decreased, which might be because adversary has more area to add adversarial perturbation. Nonetheless, the experiment results demonstrate that \detector~is still effective against all attacks for larger images. 

\begin{table}[t]
  \caption{Comparison of \detector~performance with 64$\times$64 and 32$\times$32 images from R-ImageNet. Attack Detection Rate (ADR) evaluated with TNR=95\%.}
  \label{tab:detect_large}
  \begin{center}
  \begin{tabular}{|l||c|c|c|c|}
\hline
      \multirow{2}{*}{Attack} & \multicolumn{2}{c|}{64$\times$64 images} & \multicolumn{2}{c|}{32$\times$32 images}\\
      & ~ADR~ & AUROC & ~ADR~ & AUROC \\
      %&Classes&Samples&& (bits)\\\hline
      \hline\hline
    deepFool & 33 & 79 & \textbf{42} & \textbf{87}\\
    
    CW & 37 & 76 & \textbf{70} & \textbf{91} \\
      \hline  
    PGD-4 & \textbf{86}  & \textbf{94} & 42 & 79\\
    PGD-8 & \textbf{99}  & \textbf{99} & 64 & 87\\
    PGD-16 & \textbf{99}  & \textbf{99} & 84 & 95\\
    \hline
    MIM-4 & \textbf{84}  & \textbf{93} & 51 & 79\\
    MIM-8 & \textbf{99}  & \textbf{99} & 71 & 88\\
    MIM-16 & \textbf{99}  & \textbf{99} &72 & 89\\
    \hline
    FGSM-4 & 38  & 73 & \textbf{48} & \textbf{79}\\
    FGSM-8 & 50  & 81 & \textbf{75} & \textbf{92}\\
    FGSM-16 & 57  & 85 & \textbf{89} & \textbf{96}\\
    \hline\hline
    WB & \textbf{70}  & \textbf{86} & 45 & 71\\
    \hline
  \end{tabular}
  \end{center}
\end{table}

%In Section \ref{sec:exp}, we provide empirical bounds of our approach by using three different datasets. 

%% file: relatedwork.tex
{
\section{Related Work}\label{sec:rel}
In the evasion attacks against deep neural networks, carefully crafted adversarial samples are generated to trick a deep learning model to misclassify an image/object into an incorrect class, e.g., \cite{rezaei2019target,wu2020towards,biggio2013evasion,papernot2016limitation,carlini2017towards,xiao2018generating}. Research has also shown that adversarial examples might fool time-limited humans \cite{elsayed2018adversarial}. Several adversarial detectors~\cite{Hendrycks_2016_detection,metzen2017detecting,samangouei2018defensegan} have been proposed to defend against adversarial attacks. These methods also reveal some unique detection challenges~\cite{carlini2017adversarial} e.g., the introduction of perceptually small perturbations and white-box attacks. Here, we briefly review the adversarial sample detection approaches in the literature to shed light on the design of effective and generalizable detectors.
%However, the work that has been done in the past few years gives better understanding of the problem and also lot of hope that solution against adversarial example is not too far. 

\vspace{1mm}\noindent\textbf{Separate Classifier Or Statistical Tests.} The earlier approaches to detect adversarial examples used a separately-trained classifier~ \cite{metzen2017detecting,grosse2017statistical,gong2017adversarial} or statistical properties~ \cite{Hendrycks_2016_detection,feinman2017detecting,grosse2017statistical}. However, many of these approaches were subsequently shown to be weak~\cite{carlini2017adversarial,athalye2018obfuscated}. The most recent work in detecting adversarial examples based on statistical tests \cite{Roth2019_StatTest} achieved a 99\% true positive rate (TPR) on CIFAR-10~\cite{cifar} but it was fully bypassed by later work~\cite{Hosseini2019_ByPass}, which decreased the TPR to less than 2\%. The major limitation of these approaches is that their detections are all based on identifying anomalies in sample distributions. This makes the approach vulnerable to unknown/white-box adversaries, whose adversarial samples behave like in-distribution samples. 

\vspace{1mm}\noindent\textbf{Generative Models.} The generative models have been used in adversarial image detection, e.g., \cite{lee2018simple,zheng2018robust,song2018pixeldefend,samangouei2018defensegan}. In \detector, the generative models are used in a completely different manner than the literature. Other approaches rely on likelihood density estimation through generative model, to differentiate adversarial samples from benign. On the contrary, \detector~utilizes sample generation from generative model. Performance comparison between \detector~and SOTA generative-model-based adversarial detectors are reported in Section \ref{sec:exp}, while detailed analysis of the performance differences are presented in Section \ref{sec:expanalysis}. 
%The class conditional distribution of representation layer $P(h(x)|y)$ has been used to identify adversarial samples~\cite{lee2018simple,zheng2018robust}, since the representation of each sample in one class resides in close vicinity to form a group, but the representation of the adversarial samples does not belong to any group. This approach is effective in identifying attacks that add small perturbations but cannot guarantee similar performance with more aggressive and white-box attacks. PixelDefend (PD)~\cite{song2018pixeldefend} on the other hand, uses the distribution of input samples $P(x)$ to identify adversarial samples. Like other approaches~\cite{metzen2017detecting,samangouei2018defensegan} that use input images for detection, PD is ineffective when the added perturbation is small. Overall, the generative models used in detection are ineffective against white-box attacks since the adversary can ensure high likelihood distribution for the adversarial samples. The major difference in our approach is that we use generative models for view generation instead of likelihood estimation to incorporate label information effectively.

\vspace{1mm}\noindent\textbf{Comparison with Training Samples.} Recent approaches that are proven to be successful against white-box attacks involve the comparison of test samples with all training samples \cite{Cohen_2020_CVPR,Yang_2021_detection}. Their major weakness is that they do not scale well with large datasets due to excessive memory and computational requirements.

\vspace{1mm}\noindent\textbf{Image Generation.}
Most similar to \detector~is a recently proposed image-generation-based approach~\cite{Qin2020Detecting}. The major difference between \detector~and \cite{Qin2020Detecting} is that they reconstruct the input from a high-level representation generated from a capsule classification model. But the same reconstruction technique does not give a similar performance for other classification models. \detector, on the other hand, uses an auto-regressive generation scheme that in our knowledge has never been explored in adversarial image detection before. Moreover, we use a novel multi-view approach to amplify and identify the image-label discrepancies that exist in \textit{all} successful adversarial sample attack. As a result, \detector~is not dependent on a classification or attack model, which makes it generally applicable. 

%In \cite{vacanti2020adversarial}, they have used variational auto encoder for view generation, but they only used representation layer output to introduce inconsistency, In our approach we have used classified label and input distribution to introduce inconsistency. Also, our method gives better performance on comlex dataset like cifar \cite{cifar}. 

% \subsection{Auto Regressive Image 
% Generative Models}

% ...... 

}

%% file: conclusion.tex
\section{Conclusion} \label{sec:con}
In this paper, we present an adversarial image detector named \detector. We have observed that an adversarial instance, regardless of the attack method or the level of perturbation, always contains two inherently contradictory ``souls'': the visually unchanged content and the invisible perturbation, which correspond to the true and adversarial labels, respectively. We employ a generative model to construct views to amplify the inherent discrepancies, and then design an adversary detector based on multi-view inconsistencies. Through extensive experiments, we show that \detector~achieves an average attack detection rate of 80.7\% (at 0.95 TNR) and AUROC score of 0.934 against six representative attacks, including the low-perturbation and white-box attacks. \detector~also significantly outperforms existing adversarial example detectors in both detection accuracy and robustness. Last, \detector~is a stand-alone detector that does not utilize any prior knowledge on the attacks or makes any assumption about the attack method. It could be easily adapted to any classifier/ dataset, including the ones that are already deployed. 

%Easy adoption. 